\documentclass[pdflatex,sn-mathphys-num]{sn-jnl}

\usepackage{mathrsfs}
\usepackage{graphicx}
\usepackage{multirow}%
\usepackage{amsmath,amssymb,amsfonts}%
\usepackage{amsthm}%
\usepackage{mathrsfs}%
\usepackage[title]{appendix}%
\usepackage{xcolor}%
\usepackage{textcomp}%
\usepackage{manyfoot}%
\usepackage{booktabs}%
\usepackage{algorithm}%
\usepackage{algorithmicx}%
\usepackage{algpseudocode}%
\usepackage{listings}%
\usepackage{natbib}
 \usepackage{booktabs}
\usepackage{tabularx}
\usepackage{float}  

\theoremstyle{thmstyleone}%
\theoremstyle{thmstyletwo}%

\theoremstyle{thmstylethree}%

\begin{document}

\title[Enriched Functional Tree-Based Classifiers...]{Enriched Functional Tree-Based Classifiers: A Novel Approach Leveraging Derivatives and Geometric Features}


\author*[1]{\fnm{Fabrizio} \sur{Maturo}}\email{fabrizio.maturo@unimercatorum.it}
\equalcont{These authors contributed equally to this work.}

\author[2]{\fnm{Annamaria} \sur{Porreca}}\email{annamaria.porreca@uniroma5.it}

\affil[1]{\orgdiv{Department of Economics, Statistics and Business}, \orgname{Universitas Mercatorum}, \orgaddress{\city{Rome}, \country{Italy}}}

\affil[2]{\orgdiv{Department for the Promotion of Human Science and Quality of Life}, \orgname{San Raffaele University}, \orgaddress{\city{Rome}, \country{Italy}}}


\abstract{The positioning of this research falls within the scalar-on-function classification literature, a field of significant interest across various domains, particularly in statistics, mathematics, and computer science. This study introduces an advanced methodology for supervised classification by integrating Functional Data Analysis (FDA) with tree-based ensemble techniques for classifying high-dimensional time series. The proposed framework, Enriched Functional Tree-Based Classifiers (EFTCs), leverages derivative and geometric features, benefiting from the diversity inherent in ensemble methods to further enhance predictive performance and reduce variance. While our approach has been tested on the enrichment of Functional Classification Trees (FCTs), Functional K-NN (FKNN), Functional Random Forest (FRF), Functional XGBoost (FXGB), and Functional LightGBM (FLGBM), it could be extended to other tree-based and non-tree-based classifiers, with appropriate considerations emerging from this investigation. Through extensive experimental evaluations on seven real-world datasets and six simulated scenarios, this proposal demonstrates fascinating improvements over traditional approaches, providing new insights into the application of FDA in complex, high-dimensional learning problems.}


\maketitle

\section{Introduction}
\label{sec1:intro}

In today's world, data is collected from diverse sources such as biomedical devices, smartphones, and environmental sensors, and used across applications in healthcare, environmental monitoring, and more. Technological advancements have improved our capacity to store and process this data, but managing high-dimensional datasets remains challenging. Dimensionality reduction and classification techniques are essential for effectively handling such data in medicine, environmental monitoring, security, and robotics. Key issues include irregularly spaced time points, computational complexity, the bias-variance trade-off, and the need for interpretable models with strong performance metrics such as accuracy, precision, and recall. 

In the supervised classification literature, one of the most well-known challenges is the curse of dimensionality, which arises when dealing with many variables or, in the context of time series, when there are many time points. This issue impacts numerous statistical aspects, such as distance measures, identifying causal relationships, or finding the best-performing model when many models with similar performance exist but rely on different variables. It also introduces problems like data sparsity and multicollinearity. For these reasons, the challenge of both supervised and unsupervised classification in high-dimensional data, whether it involves numerous different variables or many time points for the same variable, remains a complex and relevant area of research in mathematics, statistics, and computer science. 
Functional data analysis (FDA) is a research area that has actively tackled many of these challenges over the past decades. In FDA, dimensionality reduction is inherent, as it is achievable simply by representing the data itself.
More generally, FDA represents a statistical domain focused on the theory and application of statistical methods in scenarios where data can be expressed as functions, contrasting with the traditional representation using real numbers or vectors. FDA introduces a paradigm shift in statistical concepts, representations, modeling, and predictive techniques by treating functions as single entities. The benefits of employing FDA have been extensively discussed in contemporary literature, including the utilization of derivatives when they provide more insight than the original functions due to the nature of the phenomena, the adoption of non-parametric strategies without restrictive assumptions, data dimensionality reduction, and the exploitation of critical sources of pattern and variation \citep{Ramsay2002, Ferraty2003, Ramsay2005, Ferraty2011, Cuevas2014, Febrero2012}. 

The literature on FDA is currently dynamic and highly relevant, especially in regression, ANOVA, unsupervised classification, supervised classification, and outlier detection. Within this broad framework, we focus on supervised classification with functional predictors and a scalar response variable \citep{Preda_2007, maturo2022augmented, Aguilera2013, Cuevas2014b, Escabias2014}.
Recent research has explored the development of classification methods that combine the strengths of FDA and tree-based techniques. \citep{Yu_1999} advocated using spline trees for functional data, applying them to analyse time-of-day patterns for international call customers. Assessing variable importance in the fusion of FDA and tree-based methods was the focus of the work by \citep{Gregorutti_2015}. \citep{Moller2016} proposed a classification approach based on random forests for functional covariates. Investigating the construction of a classifier for dose-response predictions involving curve outcomes was the aim of \citep{Rahman_2019}. \citep{maturo2023supervised} proposed using functional principal components to train a classification tree. \citep{maturo2022combining} suggested combining clustering and supervised tree-based classification to enhance prediction model accuracy. Finally, \citep{maturo2022pooling} proposed an innovative evaluation of leaves' quality for functional classification trees applied to biomedical signals with binary outcomes. 
\citep{moindjie2024classification} explore methods for classifying multivariate functional data  adapting and extending PLS techniques to handle the complexity of functional data across varying domains.
\citep{brault2024mixture}  propose a mixture-based segmentation approach for heterogeneous functional data, aimed at identifying hidden structures and subgroups within complex functional datasets by combining multiple segmentations.
Recently, \citep{riccio2024supervised} proposed to exploit functional representation to increase diversity in ensemble methods and improve the accuracy of classifiers. Finally, \citep{riccio2024randomized}  suggested a new algorithm to exploit the previous idea but further improving the accuracy and variance of estimates. 
Building on the established foundation of combining FDA and statistical learning techniques, significant exploration is still needed to handle large datasets and interpret results from both statistical and causal perspectives. Research in this area is rapidly evolving and holds great potential. We expect a growing focus on improving functional classifiers' precision, interpretability, and explainability in the coming years.

Leveraging this landscape and its vast research opportunities, this paper introduces a novel functional supervised classification framework, namely the Enriched Functional Tree-Based Classifiers (EFTCs). To address the challenge of learning from high-dimensional data and enhancing functional classification performance by leveraging additional characteristics of the original data,
derivatives, curvature, radius of curvature, and elasticity are used to enrich the information provided to functional classification tree ensembles. In other words, we refer to EFTCs to denote the joint utilisation of sequential transformations for extracting unexplored features from the original signals. In essence, this approach involves viewing functions from diverse perspectives to capture additional aspects that can contribute to enhancing classification performance. Practically, it is like using a magnifying glass to reveal attributes that the original functions may miss.  Moreover, the motivation behind this proposal is also driven by the well-known fact that ensemble methods, such as tree-based classifiers, benefit significantly from introducing diversity, as it tends to improve generalisation and performance. By enriching the feature space with diverse functional characteristics, ETBCs can leverage this diversity to enhance classification accuracy further, exploiting the strengths of each transformation to capture complementary information from the data.

The paper conducts extensive experimental evaluations on seven real-world datasets and six simulated signals to measure the proposed methodology's efficacy. Comparative analyses with existing methods reveal promising results in terms of classification performance.
The study yields promising results, indicating that the enrichment approach significantly improves performance with certain methods. While our approach has been tested on functional classification trees, KNN, random forest, XGBoost, and LightGBM. However, it can be extended to other tree-based or non-tree-based classifiers, with appropriate adjustments based on our findings. This framework demonstrates notable improvements over traditional methods, offering valuable insights into applying FDA in complex, high-dimensional learning problems.

The paper's structure is as follows: Section 2 introduces the core concepts of FDA, Enriched Functional Data, and the Enriched Functional Classification frameworks, including trees, random forests, XGBoost, and LightGBM. Section 3 covers applying the proposed methods to real and simulated data.
Section 4 discusses key issues related to model explainability. 
Finally, Section 5 concludes the paper by discussing the main findings and highlighting directions for future research.

\section{Material and methods}
\label{sec2:matmed}

\subsection{Data Representation in the Functional Data Analysis (FDA) framework}
\label{sec21:matmed}

In FDA, the fundamental concept revolves around treating data functions as distinct entities. However, functional data is frequently encountered as discrete data points in practical scenarios. This means that the original function, denoted by $z = f(x)$, is transformed into a collection of discrete observations represented by $T$ pairs $(x_j, z_j)$, where $x_j$ denotes the points at which the function is assessed, and $z_j$ represents the corresponding function values at those points. We define a functional variable $X$ as a random variable with values in a functional space $\Xi$. Accordingly, a functional data set is a sample {$x_1, x_2, ..., x_N$} drawn from the functional variable $X$ \citep{Ramsay2005, Ramsay2002, Ramsay2009, Ferraty2003}.
Focusing specifically on the case of a Hilbert space with a metric $d(\cdot,\cdot)$ associated with a norm, such that $d(x_1(t), x_2 (t)) = |x_1(t) - x_2(t)|$, and where the norm $|\cdot |$ is associated with an inner product $\langle \cdot,\cdot \rangle$, such that $|x(t)|=\langle x(t),x(t) \rangle^{1/2}$, we can derive the space $\mathcal{L}^2$ of real square-integrable functions defined on $\tau$ by $\langle x_1(t),x_2(t) \rangle=\int_{\tau} x_1(t)x_2(t)\text{d}t$, where $\tau$ is a Lebesgue measurable set on $T$. Therefore, considering the specific case of $\mathcal{L}_2$, a basis function system comprises a set of linearly independent functions $\phi_j(t)$ that span the space $\mathcal{L}_2$ \citep{Ramsay2005}.

The initial step in FDA involves transforming the observed values $z_{i1}, z_{i2}, ..., z_{iT}$ for each unit $i=1,2,...,N$ into a functional form. The prevalent method for estimating functional data is basis approximation. Various basis systems can be employed depending on the characteristics of the curves.
A common approach is to represent functions using a finite set of basis functions in a fixed basis system. This can be mathematically expressed as:

\begin{equation}
x_i(t) \approx \sum_{s=1}^S c_{is}\phi_s(t),
\label{smoothfun}
\end{equation}

\noindent where, $c_i = (c_{i1}, ..., c_{iS})^T$ represents the vector of coefficients defining the linear combination, $\phi_s(t)$ is the $s$-th basis function, and $S$ is a finite subset of functions used to approximate the complete basis expansion. Another trending methodology involves leveraging a data-driven basis with Functional Principal Components (FPCs) decomposition. This approach effectively reduces dimensionality while preserving essential information from the original dataset \citep{Ramsay2005, Febrero2012}. In this context, the approximation of functional data can be expressed as follows:

\begin{equation}
x_i(t) \approx \sum_{k=1}^{K} \nu_{ik}\xi_k(t),
\label{fpca}
\end{equation}

\noindent where $K$ is the number of FPCs, $\nu_{ik}$ represents the score of the generic FPC $\xi_k$ for the generic function $x_i$ ($i=1,2,...,N$). By reducing this representation to the initial $p$ FPCs, we obtain an estimate of the sample curves, and the explained variance is given by $\sum_{k=1}^p \lambda_k$, where $\lambda_k$ denotes the variance associated with the $k$-th functional principal component. The construction of the FPCs approximation is designed such that the variance explained by the $k$-th FPC decreases with increasing values of $k$.

The variable $ T$ can be represented by various domains, such as time, space, or other parameters. The response can be categorical or numerical, leading to classification or regression challenges. However, this study is specifically concerned with a particular scenario: a scalar-on-function classification problem.
In functional classification, the objective is to forecast the class or label $Y$ for an observation $X$ within a separable metric space $(\Xi, d)$. Consequently, our methodology is tailored for functional data represented as ${y_i, x_i(t)}$, where $x_i(t)$ is a predictor curve defined for $t \in T$, and $y_i$ denotes the scalar response observed at sample $i = 1, ..., N$. The classification of a novel observation $x$ from $X$ involves the creation of a mapping $f:\Xi \longrightarrow \lbrace 0, 1, ..., C \rbrace$, referred to as a ``classifier'', which assigns $x$ to its predicted label. The error probability is quantified by $P \lbrace f(X) \neq Y \rbrace$.

\subsection{Enriched Functional Features}
\label{sec22:augfeatures}

\subsubsection{Functional Derivatives}

Let the functional derivative of order $r$ for the $i$-th curve be represented by a fixed basis system (e.g. b-splines) as:
\begin{equation}
x_i^{(r)}(t)=\sum_{j=1}^S c_{i j }^{(r)} \phi_j^{(r)}(t) \quad j=1, \ldots, S
\label{augmentedfeaturesspline}
\end{equation}

\noindent where $c_{i j }^{(r)}$ is the coefficient of the $i$-th curve, $j$-th b-spline, and $r$-th derivative order;
$\phi_j^{(r)}(t)$ is the $r$-th derivative of the $j$-th basis function. 

\citep{Ramsay2005} stressed that the selection of the basis system plays a crucial role in estimating derivatives. It is essential to ensure that the chosen basis for representing the object can accommodate the order of the derivative to be calculated. In the case of b-spline bases, this implies that the spline's order must be at least one higher than the order of the derivative under consideration. In this research, we concentrate on a b-spline basis of fourth order. 

In the following sections, we will limit our attention to the first two derivatives.
The first derivative of the function \( x_i(t) \) in the B-spline representation is given by:

\begin{equation}
x_i^{(1)}(t) = \sum_{j=1}^S c_{ij}^{(1)} \phi_j^{(1)}(t)
\end{equation}

\noindent  where \( c_{ij}^{(1)} \) are the coefficients corresponding to the first derivative of the function, and \( \phi_j^{(1)}(t) \) is the first derivative of the \( j \)-th B-spline basis function.

Similarly, the second derivative of the function \( x_i(t) \) can be expressed as:

\begin{equation}
x_i^{(2)}(t) = \sum_{j=1}^S c_{ij}^{(2)} \phi_j^{(2)}(t)
\end{equation}

\noindent where \( c_{ij}^{(2)} \) are the coefficients for the second derivative of the function, and \( \phi_j^{(2)}(t) \) is the second derivative of the \( j \)-th B-spline basis function.

In the context of functional supervised classification, B-spline versions of derivatives enhance the representation of functional features in the data by providing additional information on local variations of the curves, such as local speed and acceleration, which can be crucial for distinguishing between different classes.  In supervised classification, the speed at which a functional signal changes over time can be a key factor for class separation. For example, knowing how the heart rate varies over time in a medical dataset becomes more illuminating when considering the speed of these changes at different time intervals. 
On the other hand,  acceleration can indicate specific events or sharp changes that help differentiate one class from another, thus further improving the accuracy of the model. 
Additionally, B-spline derivatives allow for smoothed and stable derivative calculations that are less noise-sensitive than directly computed derivatives. This enriches the feature set used by classification models, improving predictive performance and class recognition.

\subsubsection{Functional Curvature and Radius of Curvature}

The curvature \( \kappa_i(t) \) of the function \( x_i(t) \) is a measure of how rapidly the function changes direction at each point \( t \). The curvature is defined as:

\begin{equation}
\kappa_i(t) = \frac{|x_i^{(2)}(t)|}{\left(1 + \left(x_i^{(1)}(t)\right)^2\right)^{3/2}}
\end{equation}

\noindent where \( x_i^{(1)}(t) \) is the first derivative and \( x_i^{(2)}(t) \) is the second derivative of the function \( x_i(t) \). 

The numerator \( |x_i^{(2)}(t)| \) represents the magnitude of the acceleration, while the denominator adjusts for the influence of the slope to ensure that the curvature is independent of the scale of \( t \). This expression provides a comprehensive measure of the function's tendency to bend at each point \( t \), capturing both the speed of change and the rate at which this speed itself changes.

The curvature \( \kappa_i(t) \) of the function \( x_i(t) \) can also be defined in terms of B-spline basis as follows:

\begin{equation}
\kappa_i(t) = \frac{\left|\sum_{j=1}^S c_{ij}^{(2)} \phi_j^{(2)}(t)\right|}{\left(1 + \left(\sum_{j=1}^S c_{ij}^{(1)} \phi_j^{(1)}(t)\right)^2\right)^{3/2}}
\end{equation}

\noindent where \( c_{ij}^{(1)} \) and \( c_{ij}^{(2)} \) are the coefficients corresponding to the first and second derivatives of the function \( x_i(t) \), and \( \phi_j^{(1)}(t) \) and \( \phi_j^{(2)}(t) \) are the first and second derivatives of the \( j \)-th B-spline basis function, respectively.

To use the curvature \( \kappa_i(t) \) in a classifier, we must extract the curvature coefficients associated with the B-spline basis functions. However, directly extracting coefficients from the above nonlinear expression for curvature is challenging because it involves a nonlinear combination of the B-spline basis functions. To overcome this, we can use the following steps for practical classification.
First, we compute the curvature \( \kappa_i(t) \) at a set of sampled points \( t_1, t_2, \dots, t_M \) over the domain \( \tau \). This results in a vector of curvature values \( \kappa_i(t_1), \kappa_i(t_2), \dots, \kappa_i(t_M) \).
Next, we fit these discrete curvature values to a B-spline basis:

\begin{equation}
\kappa_i(t) \approx \sum_{k=1}^S d_{ik} \phi_j(t)
\end{equation}

\noindent where \( \phi_j(t) \) are the B-spline basis functions, and \( d_{ik} \) are the coefficients representing the curvature in the B-spline basis.
The coefficients \( d_{ik} \) extracted from the B-spline fit are then used as features in the classifier.

The radius of curvature \( R_i(t) \) is defined as the reciprocal of the curvature \( \kappa(t) \):

\begin{equation}
R_i(t) = \frac{1}{\kappa_i(t)} \approx \frac{1}{\sum_{k=1}^S d_{ik} \phi_j(t)}
\end{equation}

In this form, the radius of curvature is represented as the reciprocal of the B-spline expansion of curvature. However, this form is not linear, which complicates direct coefficient extraction.
To facilitate coefficient extraction, we can compute the radius of curvature at sampled points and then refit these values using a B-spline basis:

\begin{equation}
R_i(t) \approx \sum_{k=1}^S e_{ik} \phi_j(t)
\end{equation}

\noindent where \( \phi_j(t) \) are the B-spline basis functions, and \( e_{ik} \) are the coefficients representing the radius of curvature in the B-spline basis.

\begin{figure}[htbp]
    \centering
    \includegraphics[width=0.8\textwidth]{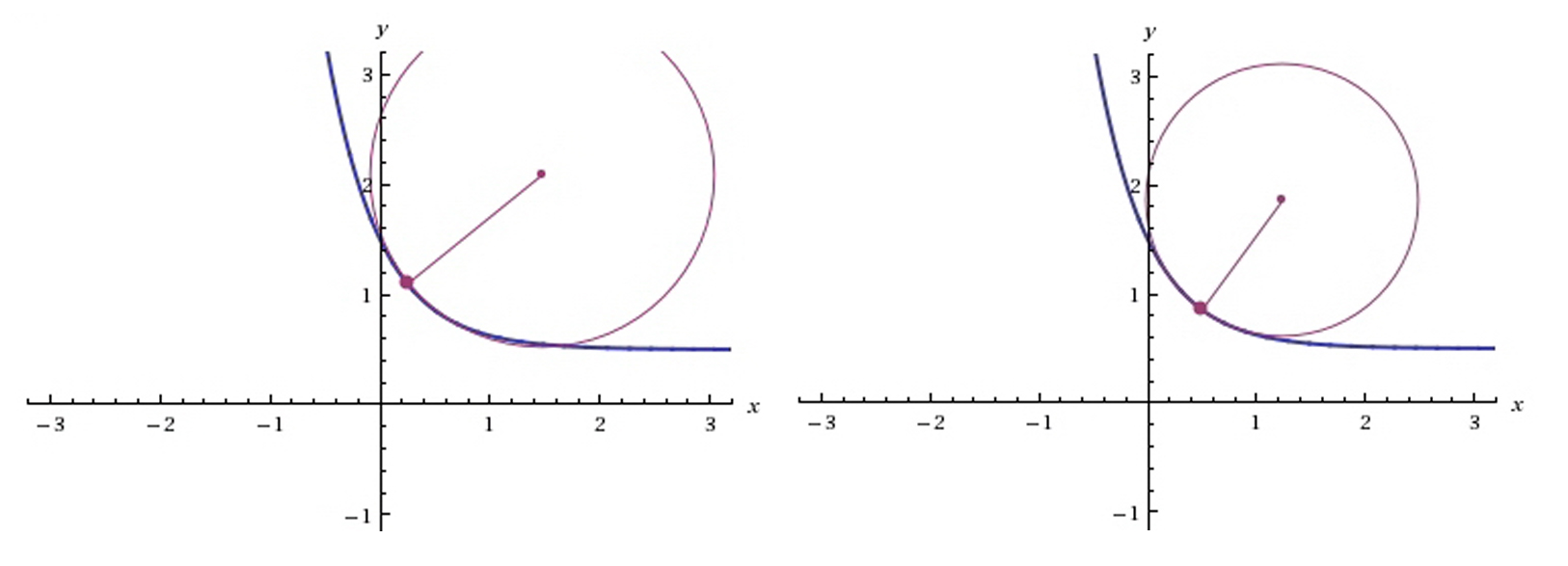}
    \caption{Curvature and radius of curvature and their geometrical interpretation.}
    \label{fig:curvature}
\end{figure}

Figure \ref{fig:curvature} illustrates two examples of the geometric interpretation of curvature and radius of curvature for a smooth curve. The blue curve represents the original function, while the purple circle is the osculating circle at a point of interest. The red dot marks the point on the curve where the curvature is being evaluated. The closeness of the osculating circle to the curve at this location visualises the curvature.

Curvature measures how sharply the curve bends at a given point, while the radius of curvature, being the reciprocal of the curvature, provides insight into the tightness or gentleness of this bend. A high curvature results in a small radius of curvature, indicating a sharp turn, whereas a low curvature corresponds to a larger radius, reflecting a gentler bend. The radius of curvature is depicted as the distance from the red dot to the center of the osculating circle.
Curvature and radius of curvature describe the local geometric properties of a curve and are valuable features for supervised classification. They provide complementary insights: curvature highlights sharp local changes in direction, making it crucial for detecting sudden transitions in the signal, while the radius of curvature offers additional context on how the curve behaves over a wider range. These features are particularly useful when distinguishing between different classes in time series or functional data.
By incorporating both curvature and radius of curvature, classification models gain a richer understanding of the signal's local and global geometry. These features capture local shape details, such as turning points or abrupt changes, which may indicate a specific class. For instance, in a medical dataset, significant variations in curvature could signal pathological conditions. Moreover, curvature and radius of curvature are robust to translations and scaling, which enhances their ability to generalize across different datasets. This robustness makes them valuable for improving classification models, as they help preserve important geometric patterns regardless of how the data is presented.

\subsubsection{Functional Elasticity}

The elasticity \( E_i(t) \) of a function \( x_i(t) \) is a measure of how responsive the function is to changes in its input, often expressed as the product of the first derivative of the function and the ratio of the input \( t \) to the function value \( x_i(t) \):

\begin{equation}
E_i(t) = x_i^{(1)}(t) \cdot \frac{t}{x_i(t)}
\end{equation}

Given that the first derivative \( x_i^{(1)}(t) \) and the function \( x_i(t) \) can be expressed using B-spline basis functions, the elasticity can be represented as:

\begin{equation}
E(t) = \left(\sum_{j=1}^S c_{ij}^{(1)} \phi_j(t)\right) \cdot \frac{t}{\sum_{j=1}^S c_{ij} \phi_j(t)}
\end{equation}

Here, \( c_{ij}^{(1)} \) are the coefficients for the first derivative, and \( c_{ij} \) are the coefficients for the original function, both represented using the same B-spline basis \( \phi_j(t) \). This expression, however, is not linear due to the ratio \( \frac{t}{x_i(t)} \), making direct coefficient extraction complex.

To facilitate the extraction of coefficients for elasticity, we can compute the elasticity at sampled points and then refit these values using a B-spline basis:

\begin{equation}
E_i(t) \approx \sum_{k=1}^S \epsilon
_{ik} \phi_k(t)
\end{equation}

\noindent where \( \phi_k(t) \) are the B-spline basis functions, and \( \epsilon_{ik} \) are the coefficients representing the elasticity in the B-spline basis.

Elasticity offers a different perspective from other geometric features, such as curvature or radius of curvature, focusing specifically on the function's rate of change relative to the input itself. While curvature deals with how sharply a function bends, elasticity quantifies the proportional change of the function in response to changes in the independent variable. This additional information can be crucial in cases where the magnitude of the input plays a role in interpreting the dynamics of the signal.

One key aspect of elasticity is its ability to capture scale-invariant properties of the function. Unlike curvature, which focuses on the geometry of the curve, elasticity reflects how the function reacts to the growth or decline of its input, making it highly relevant in scenarios where relative change matters more than absolute values. This is especially useful in fields like economics, where the proportional responsiveness of variables is more significant than their absolute variations, or in biological systems where response to stimuli may scale with intensity.

In supervised classification, elasticity highlights the signal's sensitivity to changes in the independent variable over time. For instance, in time series classification, elasticity could identify periods of rapid growth or decay that differentiate one class from another, such as distinguishing between stable versus volatile behaviour periods in financial data. Another vital aspect is elasticity's ability to reveal non-linear relationships between the input and output. Unlike derivative-based measures that are linear, elasticity incorporates both the function’s value and its rate of change, capturing a richer, non-linear interaction. Finally, elasticity complements features like curvature by focusing on input-output responsiveness, making it useful for applications requiring a holistic understanding of local behaviours and global trends. When used together in classification tasks, these features provide a more nuanced understanding of the signal’s behaviour, enriching the feature space and improving the model's ability to capture diverse patterns.

\subsubsection{The Enriched Functional Features Matrix}

Using fixed systems, the matrix of features for the original functional representation is determined by:

\begin{equation}
\mathbf{C}=
\begin{pmatrix}
    c_{11} & \dots  & c_{1S}  \\
    \vdots & \ddots   &   \vdots \\
    c_{N1} &  \dots   & c_{NS}
\end{pmatrix},
\label{featuresspline}
\end{equation}

\noindent where the generic element \( c_{is} \) is the coefficient of the \( i \)-th curve (\( i = 1,\dots,N \)) relative to the \( s \)-th (\( s = 1,\dots,S \)) basis function \( \phi_s(t) \). As a natural consequence, \( \mathbf{c}_{i} \) is the vector containing the \( i \)-th statistical unit's characteristics. 

Incorporating coefficients derived from derivatives, curvature, radius of curvature, and elasticity, we have:

\begin{enumerate}
    \item First Derivative Coefficients:
    \begin{equation}
    \mathbf{C}^{(1)}=
    \begin{pmatrix}
        c_{11}^{(1)} & \dots  & c_{1S}^{(1)}  \\
        \vdots & \ddots   &   \vdots \\
        c_{N1}^{(1)} &  \dots   & c_{NS}^{(1)}
    \end{pmatrix},
    \end{equation}
   \noindent  where \( c_{is}^{(1)} \) are the coefficients associated with the first derivative of the function.

    \item Second Derivative Coefficients:
    \begin{equation}
    \mathbf{C}^{(2)}=
    \begin{pmatrix}
        c_{11}^{(2)} & \dots  & c_{1S}^{(2)}  \\
        \vdots & \ddots   &   \vdots \\
        c_{N1}^{(2)} &  \dots   & c_{NS}^{(2)}
    \end{pmatrix},
    \end{equation}
    \noindent where \( c_{is}^{(2)} \) are the coefficients associated with the second derivative of the function.

    \item Curvature Coefficients:
    \begin{equation}
    \mathbf{D}=
    \begin{pmatrix}
        d_{11} & \dots  & d_{1S}  \\
        \vdots & \ddots   &   \vdots \\
        d_{N1} &  \dots   & d_{NS}
    \end{pmatrix},
    \end{equation}
   \noindent  where \( d_{is} \) are the coefficients representing the curvature in the B-spline basis.

    \item Radius of Curvature Coefficients:
    \begin{equation}
    \mathbf{E}=
    \begin{pmatrix}
        e_{11} & \dots  & e_{1S}  \\
        \vdots & \ddots   &   \vdots \\
        e_{N1} &  \dots   & e_{NS}
    \end{pmatrix},
    \end{equation}
   \noindent  where \( e_{is} \) are the coefficients representing the radius of curvature in the B-spline basis.

    \item Elasticity Coefficients:
    \begin{equation}
    \mathbf{F}=
    \begin{pmatrix}
        \epsilon_{11} & \dots  & \epsilon_{1S}  \\
        \vdots & \ddots   &   \vdots \\
        \epsilon_{N1} &  \dots   & \epsilon_{NS}
    \end{pmatrix},
    \end{equation}
    \noindent where \( \epsilon
_{is} \) are the coefficients representing the elasticity in the B-spline basis.
\end{enumerate}

By aggregating all the above features into a unique feature matrix $\mathbf{S}$, we can significantly enhance the power of the functional classifiers. 
It is important to emphasize that the curves in the test set are also represented using the same fixed B-spline basis system as the training set. Since the basis functions are fixed and predefined, the representation of test set curves is consistent with that of the training set. This approach ensures that the coefficients derived from the test curves are directly comparable to those obtained from the training curves. This solves a problem in the paper of \citet{maturo2022pooling}, which used the functional principal components as features.
Indeed, in a data-driven basis system, where the basis functions are derived from the data itself (e.g., using functional principal component analysis), we would face the challenge of having to project the test set curves onto a potentially different set of basis functions than those used for the training set. This could lead to inconsistencies and complications, as the basis functions might differ depending on the specific characteristics of the training data.

\subsection{Enriched Functional Classification Trees}
\label{sec:functional_classification}

In functional data analysis (FDA) context, classifying functional observations into distinct categories based on their intrinsic properties is a central problem. 
The core idea behind Enriched Functional Classification Trees (EFCTs) is to extend the classical decision tree methodology by incorporating features derived from functional data representations. This is done explicitly using B-spline coefficients of the original function and its various functional transformations, including the first and second derivatives, curvature, radius of curvature, and elasticity.

We suppose to deal with a set of functional observations \( \{ x_i(t) \}_{i=1}^N \), where each \( x_i(t) \) is a function defined over a domain \( t \in \tau \), and \( y_i \in \{1, 2, \dots, C\} \) represents the categorical outcome associated with each function.
The task of EFCTs can then be expressed as finding a mapping \( f: \mathbb{R}^{pS} \rightarrow \{1, 2, \dots, C\} \), where \( p \) is the total number of functional transformations considered (including the original function, derivatives, curvature, etc.) such that:

\begin{equation}
\hat{y}_i = f(\mathbf{S}_i) = f\left(c_{i1}, \dots, c_{iS}, c_{i1}^{(1)}, \dots, c_{iS}^{(1)}, c_{i1}^{(2)}, \dots, c_{iS}^{(2)}, d_{i1}, \dots, d_{iS}, e_{i1}, \dots, e_{iS}, \epsilon_{i1}, \dots, \epsilon_{iS}\right)^\top
\end{equation}

\noindent where \( \hat{y}_i \) is the predicted class label for the \( i \)-th observation and $\mathbf{S_i}$ is the  scores' vector for the curve $i$, with:

\begin{itemize}
    \item \( c_{ij} \) being the B-spline coefficients for the original function \( x_i(t) \),
    \item \( c_{ij}^{(1)} \) and \( c_{ij}^{(2)} \) being the coefficients for the first and second derivatives, respectively,
    \item \( d_{ij} \), \( e_{ij} \), and \( \epsilon_{ij} \) being the coefficients corresponding to the curvature, radius of curvature, and elasticity, respectively.
\end{itemize}

The feature vector \( \mathbf{S}_i \) provides a comprehensive representation of the $i$-th functional data, capturing its various transformations and ensuring that the functional nature of the data is effectively utilized within the decision tree framework. In EFCTs, each split in the tree is based on one or more of these coefficients, allowing the tree to make decisions sensitive to specific parts of the functional domain \( \tau \) and different orders of derivatives. The EFCT algorithm is illustrated in Algorithm \ref{Algo}.

\begin{algorithm}[htbp]
\caption{Enriched Functional Classification Tree (EFCT) Algorithm}
\label{Algo}
\begin{algorithmic}[1]
\State \textbf{Input:} Training data $\{(\mathbf{S}_i, y_i)\}_{i=1}^N$, where $\mathbf{S}_i$ is the feature vector of B-spline coefficients and $y_i$ is the categorical outcome.
\State \textbf{Output:} A classification tree for predicting class labels.

\Procedure{BuildTree}{$\{(\mathbf{S}_i, y_i)\}_{i=1}^N$}
    \If {Stopping criteria are met}
        \State \textbf{Return} a terminal node with class label assigned based on the majority class in the node.
    \Else
        \State Select the best coefficient $S_{ik}$ and threshold $\theta$ based on the splitting criterion (e.g., Gini impurity, information gain).
        \State Partition the data into two subsets:
        \State \hspace{1em} Left subset: $\{(\mathbf{S}_i, y_i) \mid S_{ik} \leq \theta\}$
        \State \hspace{1em} Right subset: $\{(\mathbf{S}_i, y_i) \mid S_{ik} > \theta\}$
        \State \textbf{Recursively} apply \textsc{BuildTree} to the left and right subsets.
        \State \textbf{Return} the current node with the splitting rule and child nodes.
    \EndIf
\EndProcedure

\Procedure{Predict}{$\mathbf{S}_{\text{new}}$}
    \State Start at the root node of the tree.
    \While {current node is not a terminal node}
        \If {$S_{k}(\mathbf{S}_{\text{new}}) \leq \theta$}
            \State Move to the left child node.
        \Else
            \State Move to the right child node.
        \EndIf
    \EndWhile
    \State \textbf{Return} the class label of the terminal node.
\EndProcedure

\State \textbf{Train the tree:} \textsc{BuildTree}$(\{(\mathbf{S}_i, y_i)\}_{i=1}^N)$
\State \textbf{Make predictions:} \textsc{Predict}$(\mathbf{S}_{\text{new}})$
\end{algorithmic}
\end{algorithm}

When reasoning with a single EFCT, one can consider pruning it with classical methods such as cost-complexity pruning to prevent overfitting. The increase in available features caused by enrichment with functional transformations makes pruning necessary to prevent poor generalization ability. As in the classic case, a single EFCT, although quite accurate and easy to interpret, suffers from high variance, as illustrated in the application section. For this reason, referring to ensemble methods is increasingly advantageous.

\subsection{Enriched Ensembles Methods for Functional on Scalar Classification Problems}
\label{sec:enhanced_rf}

While ECTs provide a robust framework for functional data classification using B-spline coefficients, they can be further enriched through ensemble methods. Ensemble methods, such as Random Forests, XGBoost, and LightGBM, leverage the strengths of multiple models to improve predictive performance, reduce variance, and increase robustness.

\subsubsection{Enriched Functional Random Forests (EFRF)}

The Random Forest algorithm is a natural extension of decision trees, where multiple trees are constructed, and their predictions aggregated to produce a final classification. In the context of functional data, the Enriched Functional Random Forest (EFRF) algorithm operates by constructing a collection of EFCTs, each trained on a bootstrap sample of the functional data represented by B-spline coefficients.
Each EFCT within the EFRF is constructed by recursively splitting the feature space, where the features are the B-spline coefficients. At each node in the tree, a split is made based on one of these coefficients, which corresponds to a specific aspect of the functional data, such as the value of the original function, its first derivative, second derivative, curvature, radius of curvature, or elasticity. The threshold used for the split at each node represents a critical value of a particular B-spline coefficient that best separates the data into distinct categories. For example, a split might be based on a coefficient associated with the first derivative, indicating that the decision is influenced by the rate of change in the function at a specific point in time.
In other words, each EFCT in the EFRF is built using the same process as the EFCT but with the added randomness of selecting a subset of the B-spline coefficients at each split. This process ensures that each EFCT in the forest is slightly different, reducing the correlation between EFCTs and thereby improving the overall predictive accuracy of the ensemble. The final prediction is made by aggregating the predictions of all trees in the forest, typically through majority voting.

The structure of each EFCT can be viewed as a hierarchical sequence of decisions, starting from the root node representing the entire functional dataset and progressing down to the leaf nodes where final class decisions are made. Each path from the root to a leaf node reflects a series of rules that successively refine the classification based on different features of the functional data. For instance, a path might start with a split on a B-spline coefficient related to the original function \( x_i(t) \) and then proceed with a split on a coefficient associated with the first derivative \( x_i^{(1)}(t) \), suggesting that both the function’s value and its rate of change are crucial for classifying the data.

An essential aspect of interpreting EFRF models is evaluating variable importance, which measures how often each B-spline coefficient is used in the splits and how much those splits contribute to the model’s accuracy. This analysis helps identify which functional features are most critical in distinguishing between classes. For example, if coefficients related to the second derivative are frequently selected for splits, it indicates that the acceleration of the function plays a significant role in the classification process.
Overall, while the individual trees in the forest may be complex, the EFRF model provides a coherent framework for classification by leveraging the rich information in the functional data. The consistency of using the same number of B-spline coefficients across all trees enhances the model's explainability, allowing for a meaningful understanding of how the functional data’s various transformations contribute to the classification decisions.

\subsubsection{Enriched Functional XGBoost (EFXGB)}

XGBoost (Extreme Gradient Boosting) is an advanced and scalable implementation of gradient boosting algorithms that excels in both predictive accuracy and computational efficiency. In the context of functional data analysis, we extend XGBoost by using B-spline coefficients as features derived from functional data. This approach allows the model to capture and utilise the intricate structure inherent in functional data. The model we propose is termed Enriched Functional XGBoost (EFXGB).

Let \( \mathbf{S}_i \) represent the vector of B-spline coefficients for the \( i \)-th functional observation, which includes coefficients from the original function \( x_i(t) \), its first and second derivatives, curvature, radius of curvature, and elasticity. 
The predicted class label for the \( i \)-th observation is given by $\hat{y}_i = f(\mathbf{S}_i)$.

In EFXGB, the goal is to minimise a loss function \( \mathcal{L}(\mathbf{S}, \mathbf{y}) \) over the predictions \( \hat{\mathbf{y}} \), where \( \mathbf{y} = (y_1, \dots, y_N)^\top \) are the true labels. The model builds an ensemble of EFCTs sequentially, where each EFCT \( f_m(\mathbf{S}) \) is trained to correct the errors made by the previous trees. The prediction for the \( i \)-th observation after \( m \) EFCTs is:

\begin{equation}
\hat{y}_i^{(m)} = \sum_{k=1}^{m} \alpha_k f_k(\mathbf{S}_i)
\end{equation}

\noindent where \( \alpha_k \) are weights assigned to each tree, typically learned during training. The model iteratively updates these EFCTs by minimising the following objective function:

\begin{equation}
\mathcal{L}^{(m)} = \sum_{i=1}^{N} l(y_i, \hat{y}_i^{(m-1)} + \alpha_m f_m(\mathbf{S}_i)) + \Omega(f_m)
\label{lossfun}
\end{equation}

\noindent where \( l(\cdot) \) is a differentiable convex loss function, such as logistic loss or squared error, and \( \Omega(f_m) \) is a regularization term that penalizes the complexity of the EFCT \( f_m(\mathbf{S}) \). The regularization term \( \Omega(f_m) \) typically includes the number of leaves \( T \) in the EFCT and the \( L_2 \)-norm of the leaf weights:

\begin{equation}
\Omega(f_m) = \gamma T + \frac{1}{2} \lambda \sum_{j=1}^{T} w_j^2
\end{equation}

\noindent where \( w_j \) represents the weight assigned to leaf \( j \), \( \gamma \) controls the complexity of the model, and \( \lambda \) is the regularization parameter.
During each iteration, the model calculates the first and second-order gradients of the loss function concerning the predictions, known as the gradient \( g_i^{(m)} \) and Hessian \( h_i^{(m)} \):

\begin{equation}
g_i^{(m)} = \frac{\partial l(y_i, \hat{y}_i^{(m-1)})}{\partial \hat{y}_i^{(m-1)}}
\end{equation}

\begin{equation}
h_i^{(m)} = \frac{\partial^2 l(y_i, \hat{y}_i^{(m-1)})}{\partial \hat{y}_i^{(m-1)^2}}
\end{equation}

These gradients and Hessians are used to fit the new EFCT \( f_m(\mathbf{S}) \) by minimising a second-order approximation of the loss function. The decision rules within each EFCT are based on the B-spline coefficients, allowing the model to leverage the functional characteristics of the data throughout the boosting process.

EFXGB's flexibility in handling various loss functions and incorporating regularisation makes it particularly powerful for complex functional classification tasks. Using B-spline coefficients ensures that the functional nature of the data is preserved and leveraged at each step, resulting in a model that is both accurate and with low variance. Each EFCT adds information about the functional data, gradually refining the model's predictions by focusing on correcting the errors from previous iterations.

\subsubsection{Enriched Functional LightGBM (EFLGBM)}

In this section, we extend the Light Gradient Boosting Machine (LightGBM) to functional data classification tasks by incorporating previously enriched features, including B-spline coefficients derived from the original function, its derivatives, curvature, radius of curvature, and elasticity. This extension, termed Enriched Functional LightGBM (EFLGBM), efficiently captures the structure of functional data while leveraging LightGBM's computational advantages.

Similar to EFXGB, EFLGBM constructs an ensemble of EFCTs, to minimise the same loss function \( \mathcal{L}(\mathbf{S}, \mathbf{y}) \) defined in Equation \ref{lossfun}. The prediction for the \( i \)-th observation after \( m \) EFCTs follows the same formulation:

\begin{equation}
\hat{y}_i^{(m)} = \sum_{k=1}^{m} \alpha_k f_k(\mathbf{S}_i)
\end{equation}

\noindent where \( \alpha_k \) are the weights associated with each EFCT, as described in the EFXGB section.

A critical difference between EFLGBM and EFXGB lies in the EFCT-growing strategy. While EFXGB grows EFCT via a level-wise approach, EFLGBM employs a leaf-wise growth strategy, where at each iteration, the model splits the leaf leading to the most considerable reduction in the loss function. This approach allows EFLGBM to explore more complex EFCT, potentially capturing subtle patterns in the functional data. The objective function for EFLGBM is identical to the one used for EFXGB, with the regularisation term \( \Omega(f_m) \) controlling the model's complexity through the number of leaves \( T \) and the leaf weights \( w_j \):

\begin{equation}
\Omega(f_m) = \gamma T + \frac{1}{2} \lambda \sum_{j=1}^{T} w_j^2
\end{equation}

As in EFXGB, the model relies on first and second-order gradients \( g_i^{(m)} \) and Hessians \( h_i^{(m)} \) to fit each new EFCT, using $g_i^{(m)} = \frac{\partial l(y_i, \hat{y}_i^{(m-1)})}{\partial \hat{y}_i^{(m-1)}}$ and 
$h_i^{(m)} = \frac{\partial^2 l(y_i, \hat{y}_i^{(m-1)})}{\partial \hat{y}_i^{(m-1)^2}}$.

The leaf-wise growth of EFCTs in EFLGBM, combined with the enriched functional features, enables the model to efficiently capture complex interactions in the functional data, leading to highly accurate and low-variance classification models. By focusing on leaves with the greatest potential to reduce loss, EFLGBM balances computational efficiency with model complexity, making it a robust choice for functional data classification. EFLGBM, like EFXGB, preserves the data's functional characteristics by using B-spline coefficients as input features, ensuring that the underlying structure of the functional data is leveraged throughout the boosting process. However, its leaf-wise strategy allows it to scale more efficiently, especially in large datasets where subtle functional patterns must be detected.

\section{Applications}
\label{sec3:appreal}

In subsections \ref{sec3.1:appreal} and \ref{sec3.2:appreal}, we use seven-time series datasets from the Time Series Classification Repository \citep{tsc_repository}, covering various application domains such as electrocardiogram (ECG) signals, image analysis, and energy demand. Table \ref{table:datasets_selected} summarises the main characteristics of these datasets, including the number of training and test samples, the length of the time series, and the number of classes. Supervised classification of functional data is performed using the methods described in Section 2. While we evaluate the performance of the proposed methods across all datasets, we focus particularly on the \textit{Car} dataset to illustrate the methodology in detail. This includes the steps of data preparation, applying functional classification models, and interpreting the results. We will only present the final results for the other datasets, comparing our approach with existing methods in the literature.

\begin{table}[htbp]
\centering
\renewcommand{\arraystretch}{1.3} 
\setlength{\tabcolsep}{12pt} 
\begin{tabular}{l c c c c l}
\hline
\hline
\textbf{Dataset} & \textbf{Train Size} & \textbf{Test Size} & \textbf{Length} & \textbf{No. of Classes} & \textbf{Type} \\ 
\hline
\textit{Car} & 60 & 60 & 577 & 4 & Sensor \\
\textit{ECG200} & 100 & 100 & 96 & 2 & ECG \\
\textit{ECGFiveDays} & 23 & 861 & 136 & 2 & ECG \\
\textit{ItalyPowerDemand} & 67 & 1029 & 24 & 2 & Sensor \\
\textit{Plane} & 105 & 105 & 144 & 7 & Sensor \\
\textit{Trace} & 100 & 100 & 275 & 4 & Sensor \\
\textit{TwoLeadECG} & 23 & 1139 & 82 & 2 & ECG \\
\hline
\hline
\end{tabular}
\caption{Selected Time Series Classification Datasets \citep{tsc_repository}.}
\label{table:datasets_selected}
\end{table}

In subsection \ref{sec3.3:appreal},  we test the method on six additional simulated datasets, each with a different number of classes. A detailed description of the simulation scenarios can be found in the same subsection. Table \ref{table:simulated_datasets} summarises key details of the simulated datasets, including the number of classes, time points, and the size of the training and test datasets.

\begin{table}[htbp]
\centering
\renewcommand{\arraystretch}{1.3} 
\setlength{\tabcolsep}{12pt} 
\begin{tabular}{l c c c c c}
\hline
\hline
\textbf{Scenario} & \textbf{Train Size} & \textbf{Test Size} & \textbf{Length} & \textbf{No. of Classes} & \textbf{Type} \\ 
\hline
\textit{Scenario 1} & 100 & 100 & 50 & 2 & SIMULATED \\
\textit{Scenario 2} & 100 & 100 & 50 & 2 & SIMULATED \\
\textit{Scenario 3} & 100 & 100 & 50 & 2 & SIMULATED \\
\textit{Scenario 4} & 200 & 200 & 50 & 4 & SIMULATED \\
\textit{Scenario 5} & 150 & 150 & 50 & 3 & SIMULATED \\
\textit{Scenario 6} & 150 & 150 & 50 & 3 & SIMULATED \\
\hline
\hline
\end{tabular}
\caption{Details of the simulated datasets used for classification experiments.}
\label{table:simulated_datasets}
\end{table}

\subsection{Detailed Methodology Description using the \textit{Car} Dataset}
\label{sec3.1:appreal}

The \textit{Car} dataset contains outlines of four different types of cars, extracted from traffic videos using motion information. These images were mapped onto a 1-D time series, and the vehicles were classified into one of the four categories: sedan, pickup, minivan, or SUV. Further details about the dataset are available in the work by  \citep{Thakoor2005}.
We utilize B-spline basis functions to approximate the original curves and extract enriched features, such as derivatives, curvature, radius of curvature, and elasticity, as described in Section 2.

\begin{figure}[htbp] 
\centering 
\includegraphics[width=0.8\textwidth]{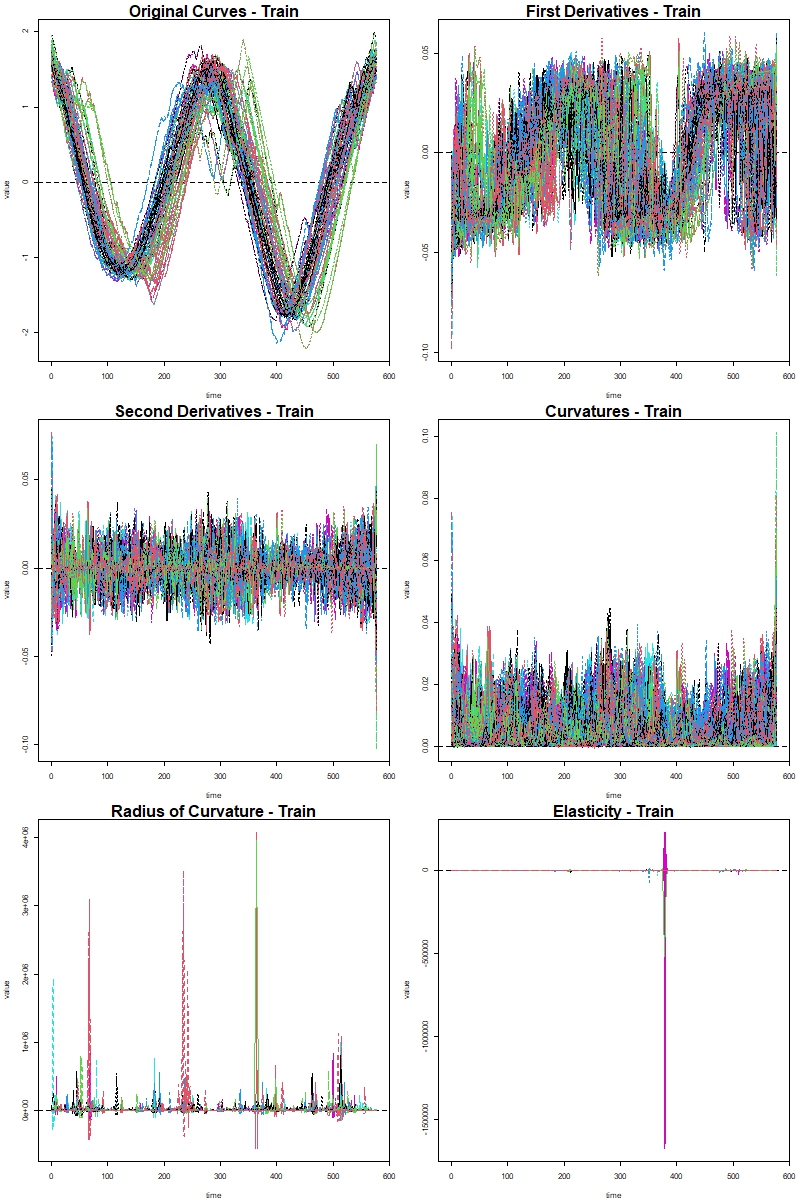} 
\caption{Functional Data for the Training Set (Car Dataset).} 
\label{cartrain} 
\end{figure}

\begin{figure}[htbp] 
\centering 
\includegraphics[width=0.8\textwidth]{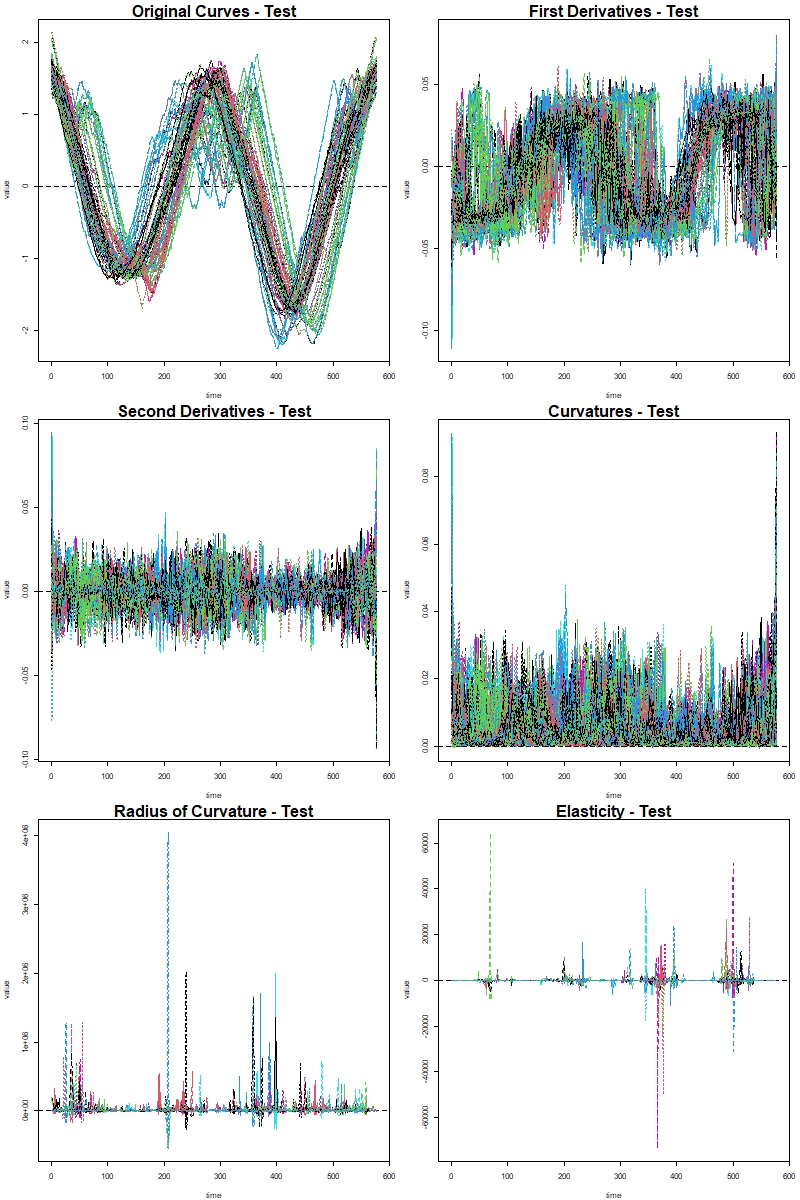} 
\caption{Functional Data for the Test Set (Car Dataset).} 
\label{cartest} 
\end{figure}

Figures \ref{cartrain} and \ref{cartest} present the functional data for the training and test sets. The original curves and the first and second derivatives, curvature, radius of curvature, and elasticity, are shown. The first plot displays the original curves, highlighting the characteristic shapes of the four vehicle types. The other plots focus on enriched functional features, such as the rate of change captured by the first derivatives, acceleration and deceleration seen in the second derivatives, the degree of bending in the curvature and radius of curvature, and the responsiveness measured by elasticity.

Concerning the functional representation through the b-splines fixed basis system, we stress that despite the fact we could select the number of bases through cross-validation on each dimension considered, in this context, we prefer to use the classic rule, i.e. the number of b-splines equals the number of time instants plus the order of b-splines minus two \citep{Ramsay2005}.
This guarantees we have an identical number of bases for each dimension. Since the coefficients of the linear combinations are the functional classifiers' features, having a different number of coefficients to represent the dimensions of each statistical unit could lead to bias within the classifiers. In other words, we avoid an imbalance between the weight of the various dimensions and give more importance to some dimensions rather than others (we aim to prevent, for example, that cross-validation recommends using hundreds of b-splines for the second derivative and few b-splines for the first derivative or other dimensions). Instead, using a consistent number of b-spline basis functions across all curve dimensions ensures that the coefficient matrices used as features have the same dimensionality. 
Once we have chosen the same number of bases for all, only based on the number of time instants and the order of the b-splines (here we always use cubic splines, therefore of order equal to 4), we proceed to extract the scores and use them as features to train different functional classifiers.

Although, from a conceptual point of view, we expect that the proposed enrichment may perform poorly when extending the enriched features to the context of functional K-NN, we want to test its performance and compare it with a traditional functional K-NN without enriched features. This choice is driven both by a desire for experimentation to understand whether performance deteriorates and to provide a comparison between the enriched tree-based classifiers and a functional classifier that, in previous studies, has often shown competitive performance compared to more advanced methods \citep{maturo2023supervised, maturo2022combining}.
The expectation that the proposed enrichment may perform poorly in functional K-NN is based on how K-NN operates as a distance-based classifier. Functional K-NN relies on calculating distances between entire functions to classify new observations based on their proximity to existing labelled data. The introduction of enriched features, such as derivatives, curvature, and elasticity, could alter the functional data's underlying geometry, leading to distorted distance metrics. Moreover, K-NN's sensitivity to local noise and outliers could further exacerbate this issue when working with enriched functional data, as the additional features might amplify minor variations in the functional curves that are irrelevant for classification.

Therefore, in this section, we adopt the Enriched Functional K-NN (EFKNN), EFCTs, EFRF, EFXGB, and EFLGBM.
The main goal is to compare the accuracy of each of these methods, using only the coefficients of the splines of the original functions (non-enriched functional classifiers) and then considering the enriched version, that is, our proposal using all the coefficients of each dimension (derivatives, curvature, radius, and elasticity).
Subsequently, to compare with other classical functional methods, which do not necessarily use splines, we refer to some known functional classifiers in the R package \textit{fda.usc} \citep{Febrero2012}.

Although parameter optimisation could improve individual model performance, we intentionally avoid in-depth optimisation for each classifier. This decision is based on two reasons. First, we compare 15 different models, each requiring separate optimisations, leading us to various configurations, even between the pairs of methods we want to directly compare (for example, FRF with and without enrichment). Hence, we aim to use the same configuration for each couple of classifiers and understand if, under the same conditions, without optimising either one or the other, the enrichment produces effects in terms of performance.
The second reason is that to produce more robust results and not limit ourselves to the trivial comparison between single accuracy values of each model, we introduce variability in the basic configurations of the hyperparameters to have a more robust comparison between accuracy distributions for each functional classifier. In practice, this randomisation we produce to compare the results turns out to be a sort of random search tuning because we can get a deeper understanding of the classifier's potential by examining the upper range of its accuracy distribution. It reveals how each model can push its performance without extensive parameter tuning. Additionally, the goal is not to achieve the best possible accuracy but to evaluate whether enriching the features improves classification and, if so, with which models it works best. Most importantly, this approach ensures that improvements or drops in performance are due to the enriched feature representations and not a result of optimised hyperparameters for any specific method. This controlled approach helps eliminate the potential confounding effect of hyperparameter tuning and creates a controlled environment to observe how including enriched functional features impacts classification accuracy directly. Therefore, while in-depth optimisation for each classifier is possible, we prioritise comparability and robustness across all methods.

Estimates' variability is introduced in several ways across different classifiers to ensure robust results that are not due to random chance or overfitting. For each classification method, randomness is injected primarily through randomly altering specific hyperparameters during each iteration and the inherent randomness in the algorithms themselves.
For the EFCTs and FCTs, the maximum depth and minimum number of samples required to split nodes are randomly selected, ensuring that different models are generated for each run. EFRF and FRF utilise their natural variability in bootstrapping and feature selection. For EFXGB, FXGB, EFLGBM, and FLGBM, parameters like tree depth, learning rate, and sample ratios are randomly adjusted in each iteration. For EFKNN and FKNN, the number of neighbours varies to observe the impact on classification accuracy. This approach ensures that the models are evaluated under a wide range of conditions, allowing a more robust performance comparison across classifiers.

In addition, several classical methods from the \texttt{fda.usc} package are used, including recursive partitioning (rpart), neural networks (nnet), Support Vector Machines (svm), Random Forest, and cross-validated elastic-net regression (cv.glmnet). In these last methods, we use the default starting parameters, introduce variability as previously to ensure robustness, and do not work on b-splines or even enriched features.
This comprehensive comparison of methods allows for a systematic evaluation of the performance of traditional and modern machine learning techniques applied to raw functional data and feature-enriched representations across all simulated scenarios.

\begin{figure}[htbp] 
\centering \includegraphics[width=0.8\textwidth]{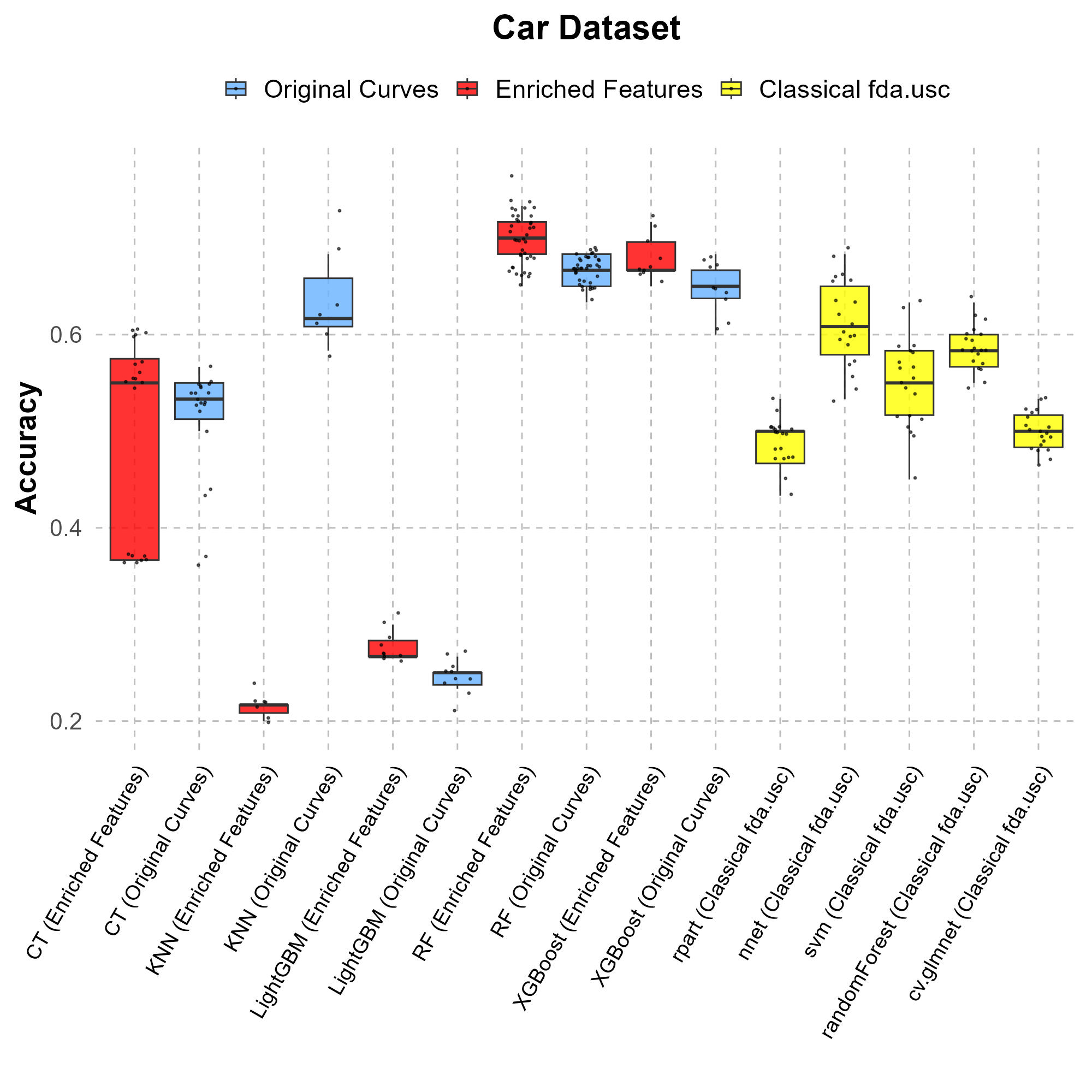} 
\caption{Comparison of Classifier Performance on the \textit{Car} Dataset. The accuracy is compared across original curves, enriched features, and classical FDA methods.} 
\label{carboxplot} 
\end{figure}

The accuracy results for different classifiers applied to the \textit{Car} dataset are summarised in Figure \ref{carboxplot}. EFRF, EFXGB, EFLGBM, and EFCTs show improvements with enriched features. Classical FDA methods implemented using the \texttt{fda.usc} package also yields competitive results but is inferior to EFRF and EFXGB. 
As we expected, EFCTs have greater variability than ensemble methods.
As expected, EFKNN is poorly suited for handling enriched features, as its performance significantly declines when incorporated. 

The enriched features significantly enhance classification accuracy, particularly in ensemble methods. EFRF and EFXGB show notable improvements, while classical methods, particularly SVM and Random Forest without enrichment and b-splines' scores, also achieve competitive results.

\subsection{Application to other Real Data}
\label{sec3.2:appreal}

This section presents the main results of comparisons across six additional datasets, each with a different number of classes. As shown in Table \ref{table:datasets_selected}, the outcomes classes range between 2 and 7, the time series lengths range from 24 to 275 instants of time, and the training and test sets have different sizes from a minimum of 23 to a maximum of 1029 instances. In all cases, the classes are well-balanced, ensuring fair comparisons.

\begin{figure}[htbp]
    \centering
    \includegraphics[width=\textwidth]{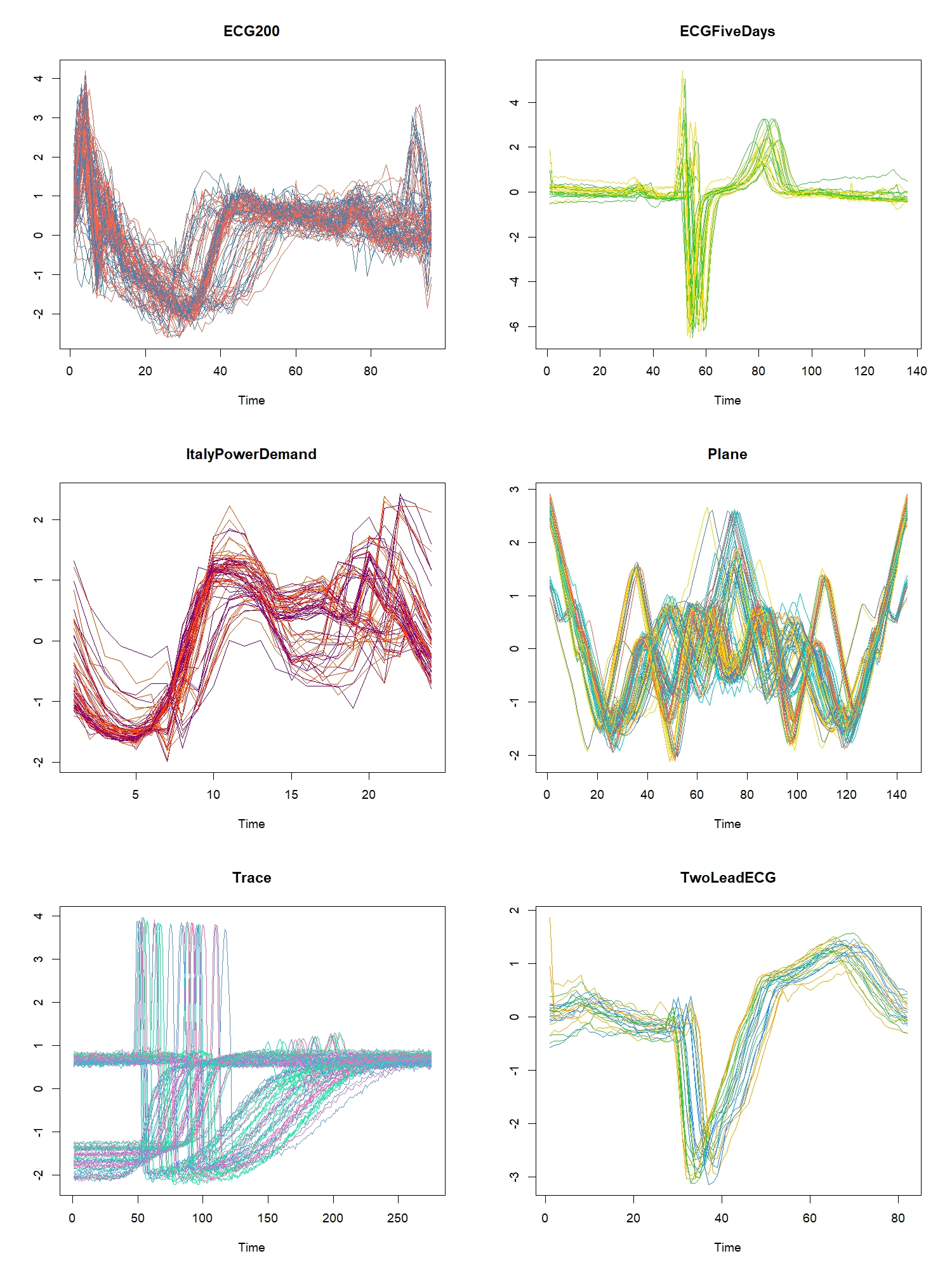}
    \caption{Original curve representations of the six datasets.}
    \label{fig:curve_representations}
\end{figure}

\begin{figure}[htbp]
    \centering
    \includegraphics[width=\textwidth]{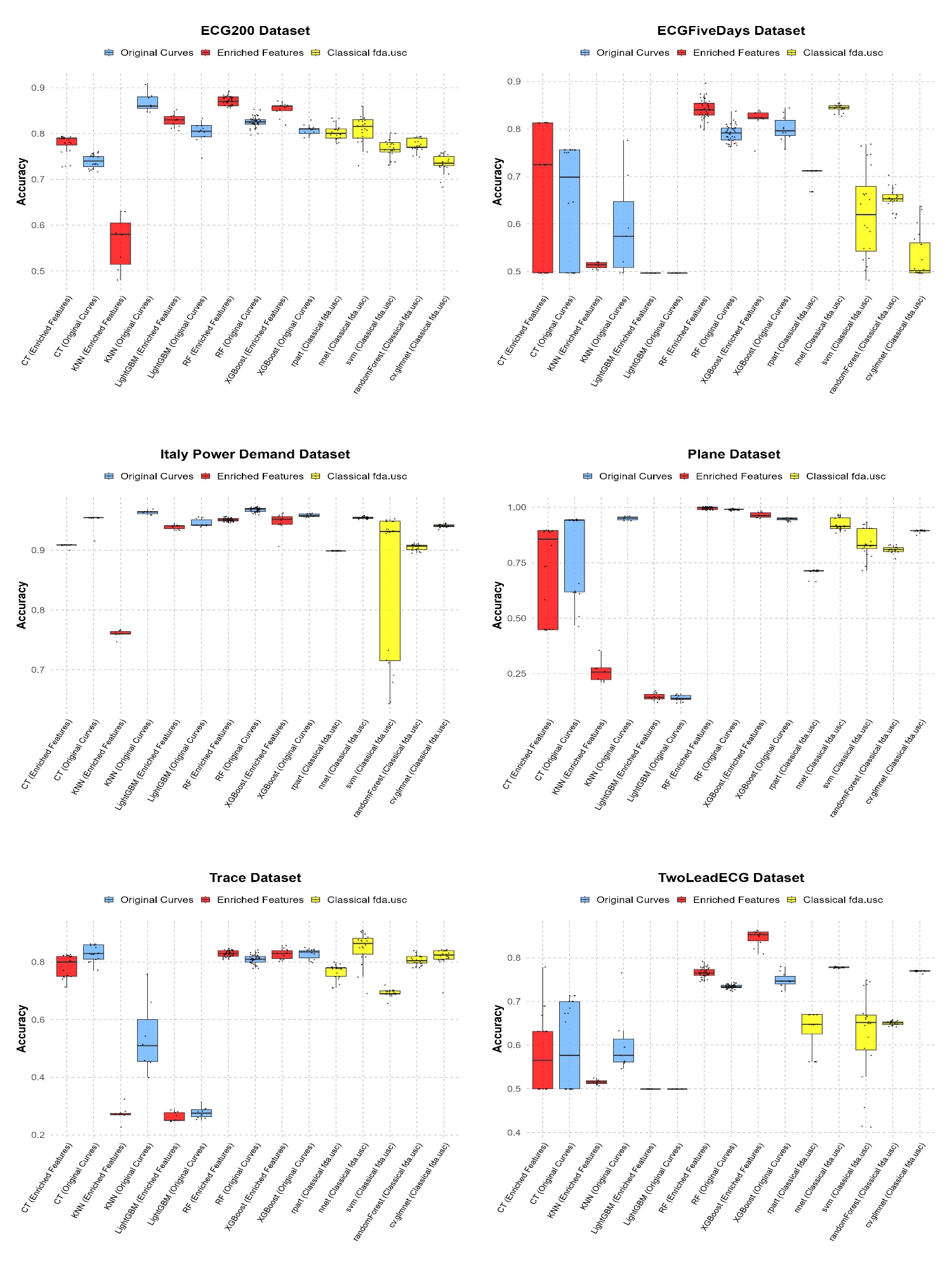}
    \caption{Comparison of accuracy across methods on six datasets.}
    \label{fig:boxplot_comparison}
\end{figure}

Figure \ref{fig:curve_representations} illustrates the original curve representations of the six datasets, showing the time series for each class. The visualisations highlight the varying characteristics of the datasets, from the simple patterns to the more intricate structures in datasets like \textit{Plane} and \textit{Trace}.
Figure \ref{fig:boxplot_comparison} shows the boxplots for accuracy across six datasets. The methods used are the same as those proposed for illustrating the \textit{Car} dataset.

Figure \ref{fig:boxplot_comparison} highlights a sharp distinction between the enriched and non-enriched versions of several classifiers for the \textit{ECG200} dataset. EFRF demonstrates the best overall performance, surpassing all other methods in accuracy. In comparison, the non-enriched FRF, FXGB, and FLGBM show consistently lower performance, reinforcing the value of enriched features. 
However, FKNN performs poorly with enriched features, showing a significant drop in accuracy compared to using original curves. Classical methods such as randomForest and cv.glmnet from the FDA.usc package perform competitively.
In the results of the \textit{ECGFiveDays} dataset, there is a clear advantage for the enriched version of the FCT, surpassing its non-enriched counterpart. On the other hand, FKNN and FLGBM perform extremely poorly with both original and enriched features, while FXGB shows an improvement in enriched and original features. The FRF demonstrates improved performance with enriched features, positioning itself as one of the top-performing models, together with \textit{nnet}. This suggests that the enriched features generally contribute positively to the performance, especially for EFRF and EFXGB.
In the \textit{Italy Power Demand} dataset, the enriched versions of FRF, FXGB, and FLGBM perform slightly worse than their original curve counterparts, making this an exception to the usual trend. FRF still delivers high accuracy and close to the best results, but the enriched features don't provide a clear advantage this time. On the other hand, the SVM from \texttt{fda.usc} shows a wide distribution of accuracy, indicating more variability in performance compared to other methods, particularly in this dataset.
In the \textit{Plane} dataset, the distinction between the enriched and original curve methods is minimal and slightly in favour of the enriched features for FRF and the two functional boosting methods. As usual, FKNN fails to perform when enriched. FCT shows significant variability and even experiences a slight drop in performance when enriched. Despite these small shifts, EFRF  remains the top-performing classifier, with consistent accuracy across the board, demonstrating its robustness even in this dataset.
In the results of the \textit{Trace} dataset, we observe a similar trend to the previous datasets. While both EFRF and EFXGB  show strong performance, the best-performing model, in this case, slightly favours \textit{nnet}. The enriched features boost the performance of FRF  compared to their original curve counterparts, but EFKNN continues to underperform. As before, FCT shows considerable variability, though still yielding relatively high accuracy with enriched and original features.
Finally, for the \textit{TwoLeadECG} dataset, FRF and FXGB show significant improvement when enriched. EFXGB  achieves the highest overall accuracy in this dataset, surpassing other classifiers. On the other hand, FKNN and FLGBM exhibit poor performance, especially in the enriched versions. This highlights the limitations of FKNN and FLGBM in handling enriched features compared to methods like FRF and FXGB, which greatly benefit from the enriched representation.

\subsection{Application to Simulated Datasets}
\label{sec3.3:appreal} 

To gather stronger evidence of the effectiveness of our methods, which already emerge from the analysis of the seven datasets presented, we conduct a simulation study. This approach allows us to evaluate performance across a broader range of scenarios, enhancing the reliability of our conclusions. Through controlled simulations, we can further assess how the enriched features interact with various classifiers and verify whether the observed improvements are consistent and significant across different synthetic data settings.

\begin{figure}[htbp]
    \centering
    \includegraphics[width=\textwidth]{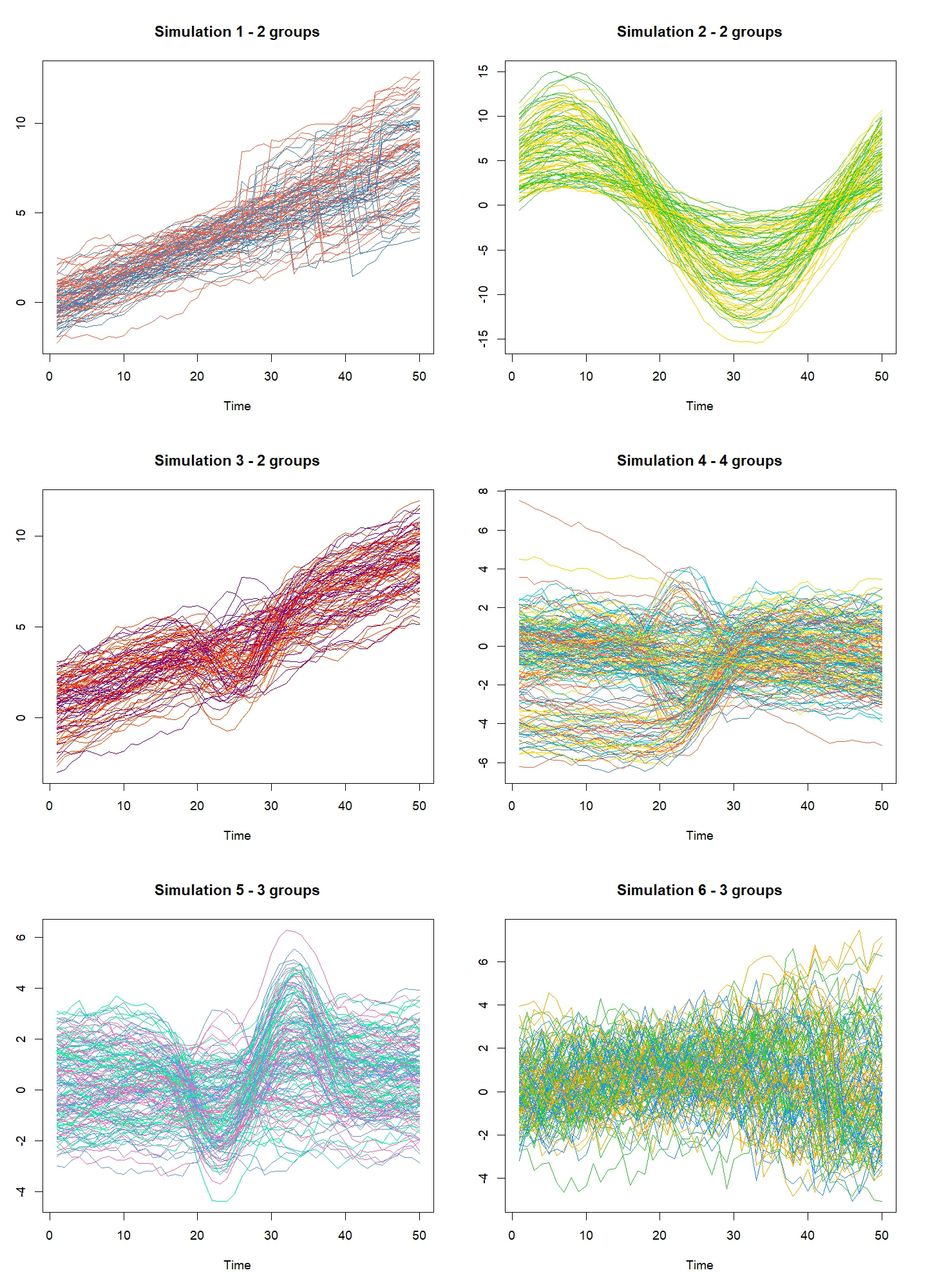}
    \caption{Simulated Scenarios with different Number of Classes.}
    \label{fig:simulazioni}
\end{figure}

To evaluate the classifier's performance, we modify and adapt several models previously considered by \citep{Cuevas_2007, Preda_2007, fdaoutlier} to generate functional data with distinct shapes. 
Figure \ref{fig:simulazioni} presents six simulated scenarios, each involving a different number of classes. The first three panels (Simulations 1, 2, and 3) represent binary classification problems, where the functional data are divided into two groups. Simulation 4 introduces a scenario with four distinct classes, while Simulations 5 and 6 involve three classes each. In binary classification scenarios, 100 curves per class are generated, and the functional data are plotted over 50 time observations. The only difference for multi-class classifications is that in these cases we generate 50 curves per class, instead of 100.

\textbf{Scenario 1:} To simulate the first scenario, we generated two groups of functional data using a model based on a Gaussian process. The base model is defined as \( X_i(t) = \mu t + e_i(t) \), where \( t \in [0, 1] \) and \( e_i(t) \) is a Gaussian process with zero mean and covariance function \( \gamma(s,t) = \alpha \exp(-\beta |t - s|^{\nu}) \). The two groups are created by adjusting some parameters in a way that introduces moderate differences between them, making the classification task non-trivial but not too simple.
For the first group, we set \( \mu = 8 \) and generated 100 curves over 50 time points. For the second group, we slightly modified the base model by introducing a shift in the function, defined as \( X_i(t) = \mu t + q k_i I_{T_i \leq t} + e_i(t) \), where \( q = 3 \) and \( k_i \in \{-1, 1\} \) with equal probability, and \( T_i \) is drawn from a uniform distribution over \( [a, b] = [0.5, 0.9] \). This shift creates functional curves for the second group that differ moderately from those in the first group, ensuring the classification task is not overly simple.
The covariance structure for both groups is controlled by the parameters \( \alpha = 1 \), \( \beta = 1 \), and \( \nu = 1 \), ensuring consistent variability across the curves. The probability of the shift being positive or negative is set to 0.5 to avoid overly distinct group separation. 

\textbf{Scenario 2:} For the second scenario, we generated two groups of periodic functional data using a model based on sinusoidal components with Gaussian noise. The base model for the first group is defined as \( X_i(t) = a_{1i} \sin(\pi t) + a_{2i} \cos(\pi t) + e_i(t) \), where \( e_i(t) \) is a Gaussian process with zero mean and covariance function \( \gamma(s,t) = \alpha \exp(-\beta |t - s|^{\nu}) \). The parameters \( a_{1i} \) and \( a_{2i} \) were set to \( a_{1i} = 1 \) and \( a_{2i} = 8 \), respectively, to generate periodic curves for the first group.
For the second group, we applied a slight variation to the model by adding a shift, resulting in the modified model \( X_i(t) = (b_{1i} \sin(\pi t) + b_{2i} \cos(\pi t)) (1 - u_i) + (c_{1i} \sin(\pi t) + c_{2i} \cos(\pi t)) u_i + e_i(t) \), where \( u_i \) follows a Bernoulli distribution with \( P(u_i = 1) = 0.60 \), and \( b_{1i} \in [1.5, 2.5] \), \( c_{1i} \in [5, 10.5] \), creating functional curves that have subtle differences from those in the first group, while still retaining the periodic nature of the data.
The covariance structure remains the same for both groups, with \( \alpha = 1 \), \( \beta = 1 \), and \( \nu = 1 \). The variations introduced by the parameters \( b_{1i} \) and \( c_{1i} \), combined with the probabilistic shift governed by \( u_i \), create a moderate difference between the two groups, ensuring that the classification problem is non-trivial.

\textbf{Scenario 3:} For the third scenario, we generated two groups of functional data using a model that introduces differences in the shape of the curves over a specific portion of the domain. The base model for the first group is defined as \( X_i(t) = \mu t + e_i(t) \), where \( e_i(t) \) is a Gaussian process with zero mean and covariance function \( \gamma(s,t) = \alpha \exp(-\beta |t - s|^{\nu}) \). For this group, we set \( \mu = 8 \).
The second group is generated by applying a shift and a change in the shape of the function, modeled as \( X_i(t) = \mu t + (-1)^u q + (-1)^{1-u} \left( \frac{1}{\sqrt{\pi r}} \right) \exp(-z(t - v)^w) + e_i(t) \), where \( u \) follows a Bernoulli distribution with probability \( P(u = 1) = 0.1 \). In this scenario, we set \( q = 1.8 \), \( r = 0.02 \), \( z = 90 \), and \( w = 2 \). The parameter \( v \) is drawn from a uniform distribution over the interval \( [0.45, 0.55] \), introducing a localized change in the shape of the curve for the second group.
The covariance structure for both groups is controlled by the parameters \( \alpha = 1 \), \( \beta = 1 \), and \( \nu = 1 \), ensuring consistent variability across the curves. The slight shift and shape changes in the second group make the classification task more challenging, as the groups are not trivially distinguishable.

\textbf{Scenario 4:} For the fourth scenario, we used the same model described in the third simulation. The model introduces differences in the shape of the curves for a portion of the domain. The base model is given by \( X_i(t) = \mu t + e_i(t) \), where \( e_i(t) \) is a Gaussian process with zero mean and covariance function \( \gamma(s,t) = \alpha \exp(-\beta |t - s|^{\nu}) \).
For the first two groups (Group 1 and Group 2), we set \( \mu = 0 \), \( q = 1 \), and controlled the timing of the shift and shape change using a uniform distribution for \( v \), drawn from the interval \( [0.45, 0.45] \). The other parameters governing the shape change were \( r = 0.02 \), \( w = 2 \), and \( z = 90 \). The covariance parameters were set to \( \alpha = 1.3 \), \( \beta = 1.2 \), and \( \nu = 1 \). 
For the second set of groups (Group 3 and Group 4), we introduced further variations. Here, we used \( \mu = -2 \), \( q = 1.8 \), and controlled the shift with \( v \) drawn from the interval \( [0.15, 0.15] \). The shape-related parameters were set to \( r = 0.01 \), \( w = 5 \), and \( z = 90 \). The covariance parameters were adjusted to \( \alpha = 0.8 \), \( \beta = 0.8 \), while keeping \( \nu = 1 \). 

\textbf{Scenario 5:} For the fifth scenario, we used the same model described in the previous simulations, but generated three distinct groups.
For Groups 1 and 2, the parameters were set as follows: \( \mu = 0 \), \( q = 1.8 \), and the timing of the shape and shift was controlled by drawing \( v \) from the interval \( [0.45, 0.45] \). The shape-related parameters were configured as \( r = 0.02 \), \( w = 2 \), and \( z = 90 \). The covariance parameters were set to \( \alpha = 1 \), \( \beta = 1 \), and \( \nu = 1 \).
For Group 3, we introduced different parameter values to create a distinct third group. Specifically, we set \( \mu = 1 \), \( q = 0.8 \), and drew \( v \) from the interval \( [0.65, 0.65] \) to control the shift. The other parameters remained the same: \( r = 0.02 \), \( w = 2 \), and \( z = 90 \), while the covariance parameters were kept at \( \alpha = 1 \), \( \beta = 1 \), and \( \nu = 1 \).

\textbf{Scenario 6:} For the sixth scenario, we adapted the model used in the first simulation to generate three distinct classes by adjusting the parameters for each class.
For the first two groups (Group 1 and Group 2), the model was configured with \( \mu = 2 \), \( q = 3 \), and a uniform distribution for \( T_i \) drawn from the interval \( [0.6, 0.75] \). The covariance parameters were set to \( \alpha = 2 \), \( \beta = 1 \), and \( \nu = 0.5 \).  
For the third group, we applied further parameter variations to create a distinct class. Here, \( \mu = 2 \) and \( q = 3 \) were kept the same, but the uniform distribution for \( T_i \) was drawn from a narrower interval \( [0.8, 0.9] \). The covariance parameters remained unchanged, with \( \alpha = 2 \), \( \beta = 1 \), and \( \nu = 0.5 \). This group was generated with 50 curves, introducing more pronounced differences compared to the first two, adding complexity to the classification task.

\begin{figure}[htbp]
    \centering
    \includegraphics[width=\textwidth]{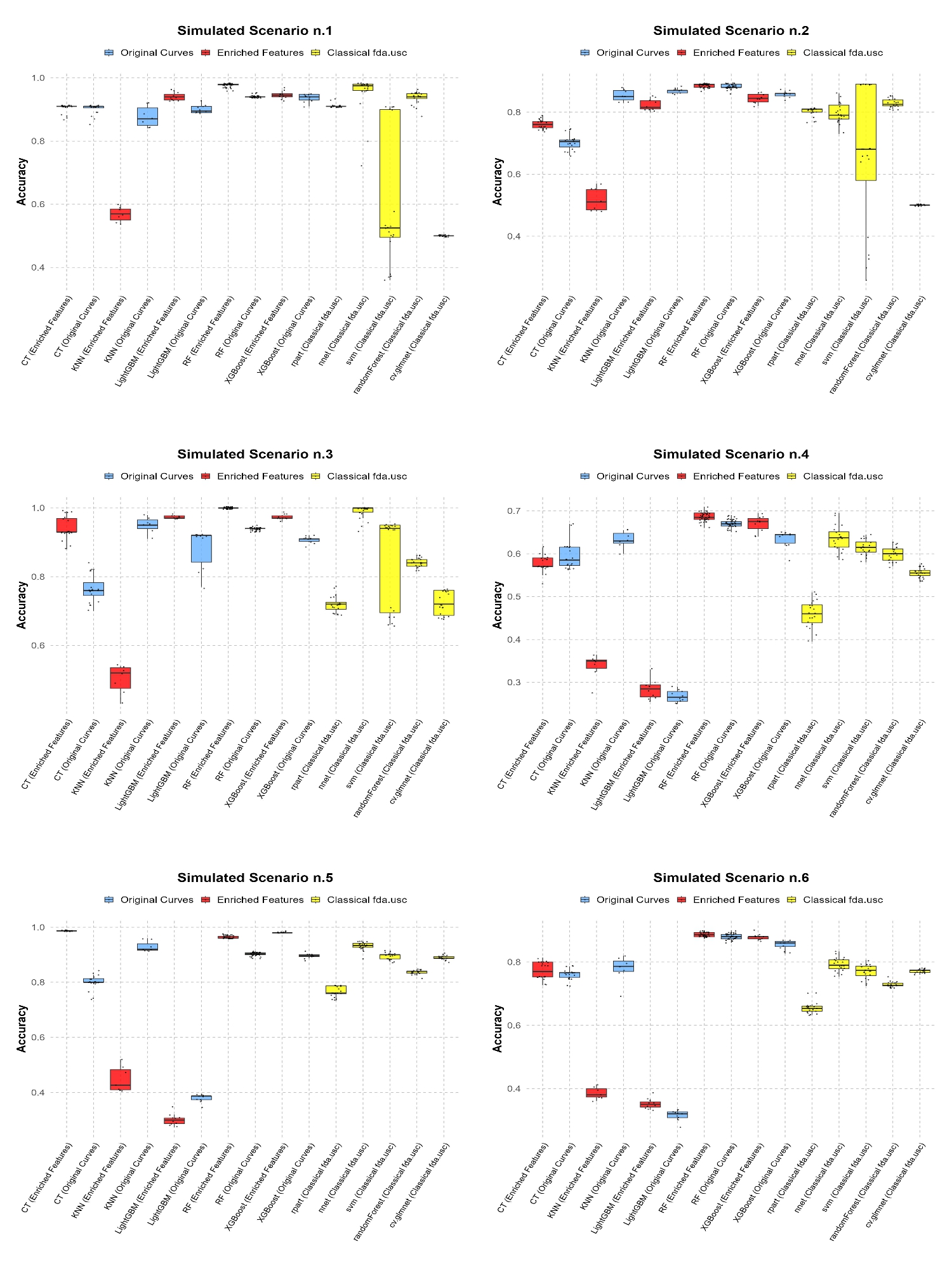}
    \caption{Simulated Scenarios Results.}
    \label{fig:simulazioni.risultati}
\end{figure}

Figure \ref{fig:simulazioni.risultati} presents the accuracy results for each of the six simulated scenarios, comparing the performance of various classification methods applied to both the original curves and enriched feature representations alongside classical functional data analysis methods.

The binary classification task in Scenario n.1 shows that the enriched features generally outperform the original curve methods. The enriched version for FCT provides better accuracy than the original curves. Similarly, for FKNN and FLGBM, the enriched feature versions yield higher accuracy than their counterparts. FRF-enriched features outperform all other methods, achieving the best overall accuracy. FXGB slightly improves with enriched features, though its performance remains behind FRF. Classical methods, such as \texttt{nnet}, perform well but are outperformed by FRF enriched, making \texttt{nnet} the second-best approach.

In Scenario 2, the enriched features yield better performance for FCT, showing clear improvement compared to the original curves. FRF, on the other hand, performs similarly for both original and enriched features, with no difference in accuracy. However, other methods such as FKNN, FLGBM, and FXGB show a slight decrease in performance when enriched features are used. Notably, FKNN with enriched features performs significantly worse than the original curve version, highlighting that this approach does not work well with FKNN. 

The binary classification problem in Scenario n.3  highlights a more robust performance from the enriched feature methods. FCT, FKNN, and FLGBM all perform better with enriched features than their original curve counterparts. FRF and FXGB, while already performing well with original curves, show slight improvements when enriched features are used. The only method that has performance comparable to enriched FRF is \texttt{nnet}.

Scenario 4 deals with a four-class problem. The enriched features improve performance for several methods. FRF with enriched features emerges as the best performer, achieving the highest overall accuracy. FLGBM and FXGB also benefit from enriched features, showing clear improvements compared to their original curve counterparts. FCT and FKNN, on the other hand, do not exhibit substantial gains from the enriched features. Among the classical methods, \texttt{nnnet} is the best.

In Scenario n. 5's three-class classification problem, FRF, FCT, and FXGB show improved performance when using enriched features. FRF remains highly competitive, but FXGB, with enriched features, emerges as the best performer. FCT also benefits from enriched features, achieving better accuracy than its original curve counterpart. However, FKNN continues to perform poorly with enriched features, reinforcing the expectation that this method is not well-suited for feature enrichment.

In the final three-class scenario n.6, enriched feature methods again tend to outperform their original curve counterparts. FCT and FLGBM show better results when using enriched features, while FRF and FXGB remain competitive with minimal differences between the two approaches.

\section{On Enriched Functional Tree-Based Classifiers Interpretability and Explainability}

There is no doubt that enriching functional features provides fascinating benefits in terms of accuracy. However, within the proposed framework, we must pay particular attention to two distinct aspects: interpretability and explainability. Although the primary focus of this paper is not on explainable artificial intelligence (XAI), but rather on introducing methodologies to enhance performance, some considerations regarding interpretability and explainability within the proposed framework can help deepen the understanding of the model and suggest potential avenues for future research.

\subsection{Enriched Functional Classification Trees' Interpretability}

From the perspective of interpretability, it is evident that, as with all ensemble methods, we lose the ability to interpret the classification rules easily. Simpler models, such as regression or classification trees, allow for a straightforward interpretation of how predictors affect the outcome, but they lose credibility when assumptions are violated in the former and due to high variability in the latter. Consequently, a model perfect for interpretability, accuracy, and low variance does not exist.

Focusing on the EFCT model, however, interpretability is still possible. The EFCT is an extension to enriched functional data of the un-enriched FCT (Functional Classification Tree) proposed by \citep{maturo2023supervised}. Therefore, with appropriate considerations for derivatives, curvature, radius of curvature, and elasticity, it is always possible to interpret the splitting rules in EFCT. Similarly, following the approach of \citep{maturo2023supervised}, it is also feasible to construct both theoretical separation curves (based on the formula that reconstructs the separation curve as a linear combination of basis functions) and empirical separation curves (based on the actual curve that is closest to the theoretical separation curve). The main difference here is that the reconstruction is based on splines rather than functional principal components, and the interpretation must be made based on the specific functional transformation involved in each node's splitting rule.
Thus, some splitting rules between groups of curves that end in a left or right node must be interpreted based on the original curves, speed, acceleration, curvature and radius of curvature, elasticity, and their intrinsic meanings.

Let \( \mathbf{S}_i = (s_{i1}, s_{i2}, \dots, s_{iD}) \) represent the B-spline coefficients of the \( i \)-th functional observation in the \( D \)-dimensional enriched feature space. Each dimension \( d \) corresponds to different aspects of the functional data, such as the original function, derivatives, curvature, radius of curvature, or elasticity.
At each node \( z \) in the classification tree (EFCT), a decision is made to split the data based on a specific feature or combination of features, which can include scores from any dimension (e.g., original functions, derivatives, or geometric features like curvature and radius of curvature). The theoretical separation curve at node \( z \), denoted by \( f_{\text{sep}, z}(t) \), is defined as a linear combination of the B-spline basis functions and the corresponding generalised coefficients selected up to that specific node.
We introduce \( \gamma_{zs} \) as the generalised coefficients representing any feature in the enriched feature matrix, whether corresponding to the original function, derivatives, or any geometric features such as curvature or elasticity.

For the \( z \)-th node, the theoretical separation curve is expressed as:

\begin{equation}
f_{\text{sep}, z}(t) = \sum_{s \in \Omega_z} \gamma_{zs} \phi_s(t)
\end{equation}

\noindent where
 \( \Omega_z = \{k_1, \dots, k_{z}\} \) is the set of B-spline basis functions and corresponding generalised coefficients selected up to the split at node \( z \),
\( \gamma_{zs} \) are the generalised coefficients associated with the \( s \)-th basis function \( \phi_s(t) \), where \( \gamma_{zs} \) generalises the coefficients for any functional transformation (original function, first/second derivatives, curvature, etc.),
 \( \phi_s(t) \) is the B-spline basis function corresponding to the feature involved in the split.

In this notation, \( \gamma_{zs} \) is the generic element of the enriched feature matrix, which we previously defined in Equation \ref{expandedfeaturematrix}. \( \gamma_{zs} \) refers to any coefficient from this expanded feature matrix, covering all possible dimensions (original functions, derivatives, and geometric features like curvature, elasticity, etc.). As each split occurs, the specific coefficients involved are determined by the splitting rule, which can span different dimensions of the feature matrix.

The interpretation of the theoretical separation curve \( f_{\text{sep}, z}(t) \) depends on the type of B-spline basis functions involved in the split, which can vary across different dimensions.
If the basis functions correspond to the original function, the separation curve reflects the overall shape of the functional signal.
If the basis functions represent first or second derivatives, the separation curve highlights the signal's local speed or acceleration.
If the split involves curvature or radius of curvature, the separation curve focuses on how sharply the signal bends at different points.
If the split is based on elasticity, the separation curve captures the function’s responsiveness to changes in its input.
Thus, the type of functional feature that drives the split dictates the specific interpretation of the theoretical separation curve. The generalised coefficient \( \gamma_{zs} \) ensures that the combination of different functional transformations is flexible and accurately represents the decision boundary at each node of the classification tree.

\subsection{Enriched Functional Tree-Based Classifiers Explainability}

Focusing on ensemble models, it is natural that interpretability is diminished, and we must instead rely on explainability as a tool to understand what is happening within the black-box model. The unique aspect of the proposed framework is that introducing these enriched features can result in correlations between variables. While it is widely accepted that multicollinearity does not pose significant issues in tree-based methods, unlike in multiple regression where it can artificially inflate R-squared values, it is essential to acknowledge that the presence of multicollinearity can distort explainability measures in black-box models. 
In other words, while from a performance standpoint, the introduction of enriched features has a purely positive effect by improving accuracy through the observation of various functional characteristics and increasing the diversity of the ensemble, which further boosts accuracy and significantly reduces variance, there is a trade-off when it comes to explainability measures. This results in a potential bias, as the complexity added by the enriched features can make it more challenging to understand the model's internal decision-making processes fully.
In other words, their importance measures may become skewed when introducing correlated scores.

In reality, while it is true that enrichment exacerbates this bias in explainability measures, the solution is relatively straightforward. The key is that when performing enrichment, as proposed in this study, it is essential to rely on variable importance measures that account for correlations between features. Despite the fixed basis system producing uncorrelated functions, the substantial increase in predictors when enrichment introduces numerous new features across many dimensions, including curvature, radius of curvature, and elasticity,  can lead to correlations between coefficients across different splines and dimensions.
To solve this issue, we can use two possible approaches. The first option is to condition on the scores of the same B-spline of different dimensions when calculating the importance of a feature. For example, when assessing the importance of the B-spline scores for the radius of curvature, we could condition on the B-spline scores for the derivatives, curvature, elasticity, and original functions, as there is likely a significant association between these coefficients. The second alternative is not to assume any correlation between the scores of the same b-splines' different dimensions and conditioning on all correlated variables beyond a certain threshold when assessing feature importance via classical methods. The latter approach is similar to those used in bioinformatics, where the number of independent variables often far exceeds the number of observations, resulting in high dimensionality \citep[see e.g.][]{fuzzyforest}. By accounting for all potential correlations, this method provides a more robust and reliable measure of model explainability. Extending the two approaches to the context described is quite immediate.

Let \( \mathbf{S}_i = (s_{i1}, s_{i2}, \dots, s_{iD}) \) represent the B-spline coefficients of the \( i \)-th functional observation in the \( D \)-dimensional enriched feature space. Each dimension corresponds to a different aspect of the functional data, such as original function coefficients, first derivatives, second derivatives, curvature, radius of curvature, and elasticity.
Let \( I(f, S_j) \) represent the importance of a feature \( S_j \) (e.g., the B-spline scores for the radius of curvature) in predicting the outcome \( y \).

The first approach can be summarized as follows. Let \( C_j \) represent the set of coefficients for these associated dimensions (i.e. associated in the sense that we deal with the scores of the same b-spline function used to reconstruct different transformations of the functional data). The conditional feature importance of \( S_j \) (e.g., radius of curvature) is given by:

\begin{equation}
I(f, S_j \mid C_j) = \mathbb{E}_{S_j \mid C_j} \left[ L(f(S_j, C_j), y) \right] - \mathbb{E}_{S_j \mid C_j} \left[ L(f(C_j), y) \right]
\label{soluz1}
\end{equation}

\noindent where
\( f(S_j, C_j) \) represents the model including both the feature \( S_j \) and the conditioning set \( C_j \),
\( f(C_j) \) represents the model excluding \( S_j \) but including \( C_j \),
\( L(f, y) \) is the loss function used to evaluate the model (e.g. cross-entropy),
\( \mathbb{E}_{S_j \mid C_j} \) denotes the expectation conditioned on \( C_j \).
Equation \ref{soluz1} quantifies the difference in performance between the full model (with \( S_j \) and \( C_j \)) and the reduced model (without \( S_j \), but conditioned on \( C_j \)). This approach provides the conditional importance of \( S_j \) by controlling for correlations with the associated dimensions.

Alternatively, the second approach does not assume any direct correlation between the scores of different dimensions. Instead, it conditions on all variables correlated beyond a certain threshold. Let \( \rho_{ij} \) denote the correlation between two feature dimensions \( S_i \) and \( S_j \). We define the set of conditioning variables \( C_j \) based on a threshold \( \tau \) as:

\begin{equation}
C_j = \{ S_i : |\rho_{ij}| > \tau \}
\end{equation}

The importance of \( S_j \) is then calculated by conditioning on the set \( C_j \) as in Equation 
\ref{soluz1} but, in this case, the conditioning set \( C_j \) includes all variables that exceed the correlation threshold \( \tau \), providing a more flexible method to account for the correlations within the enriched feature space. This approach is particularly useful when the correlations are not restricted to certain dimensions but are instead scattered across the feature space.

\section{Discussion and Conclusions}
\label{sec5:discussion}

The evolving field of supervised curve classification has made significant advances in recent decades, yet integrating Functional Data Analysis (FDA) with tree-based classifiers remains an area ripe for further development. While some previous studies have examined this combination from various perspectives, critical areas still require deeper exploration. Key areas for enhancement include improving the accuracy of functional classifiers, developing advanced graphical tools for interpreting classification rules, conducting comprehensive simulation studies, and designing effective strategies for optimising parameters in the supervised classification of functional data.

This paper introduces a novel supervised classification strategy that synergises FDA with tree-based ensembles to extract richer insights from curve analysis. The proposed Enriched Functional  Tree-Based Classifiers (EFTCs) address the challenges associated with high-dimensional data, focusing on improving classification performance. By incorporating additional features derived from various functional transformations, such as sequential derivatives, curvature, the radius of curvature, and elasticity, the enriched functional data strategy captures detailed information about the functional data's global and local behaviour. 

Extensive experimental evaluations on real-world and simulated datasets underscore the effectiveness of the proposed approach. The results demonstrate substantial improvements in classification performance over existing methods, confirming the value of the enriched functional features in managing high-dimensional data. The proposed classifiers effectively capture local characteristics often overlooked by traditional methods, highlighting the importance of these additional features in achieving accurate classification, even in scenarios involving multiple classes and complex curve shapes.
Furthermore, the enhanced performance observed in the EFTCs can also be attributed to introducing diversity, a crucial factor in ensemble methods. By incorporating multiple perspectives of the functional data, such as derivatives, curvature, and elasticity, into the model, we effectively increase the variety of decision patterns available to the ensemble, allowing it to capitalise on the complementary strengths of each feature and ultimately boost classification accuracy.
This significant result, achieved through a truly original approach to introducing diversity in the ensemble by leveraging the available functional tools, aligns with insights from the broader machine learning literature in non-functional contexts. This innovative integration strengthens the model's performance and opens new pathways for exploring ensemble methods in functional data analysis.

While this study uses B-splines for feature extraction, the underlying methodology can be extended to other functional transformations, functional classifiers, and basis functions. The fixed-basis system here offers a consistent framework for training and testing, avoiding the complications associated with data-driven basis systems, where the basis functions may vary between datasets.
Future research could enhance the interpretability and explainability of these models, or explore the integration of a weighted selection of the number of splines, potentially guided by cross-validation criteria, a choice deliberately avoided in this context, as explained in Section 2.
At the same time, once it has been demonstrated that the enrichment performs well in terms of accuracy, future studies can further explore parameter optimisation by focusing on specific functional classifiers. As thoroughly explained in Section 2, we avoided deep optimisation to maintain an experimental setup conducive to comparison, which was the study's primary objective.

\subsection*{Funding and/or Conflicts of Interest/Competing Interests}

The authors confirm that they received no support from any organization for the submitted work. They also declare no affiliations or involvement with any organization or entity with a financial or non-financial interest in the subject matter of this manuscript.

\subsection*{Use of Generative AI in Scientific Writing}

AI-assisted technologies were used only in the writing process to improve the readability and language of the manuscript. The authors reviewed and edited the content as necessary and took full responsibility for the publication's content.

\subsection*{Data availability statement}
 
The authors used publicly available data for real-world applications. Simulation data can be provided upon request.

\bibliography{_biblio}


\begin{thebibliography}{28}
\ifx \bisbn   \undefined \def \bisbn  #1{ISBN #1}\fi
\ifx \binits  \undefined \def \binits#1{#1}\fi
\ifx \bauthor  \undefined \def \bauthor#1{#1}\fi
\ifx \batitle  \undefined \def \batitle#1{#1}\fi
\ifx \bjtitle  \undefined \def \bjtitle#1{#1}\fi
\ifx \bvolume  \undefined \def \bvolume#1{\textbf{#1}}\fi
\ifx \byear  \undefined \def \byear#1{#1}\fi
\ifx \bissue  \undefined \def \bissue#1{#1}\fi
\ifx \bfpage  \undefined \def \bfpage#1{#1}\fi
\ifx \blpage  \undefined \def \blpage #1{#1}\fi
\ifx \burl  \undefined \def \burl#1{\textsf{#1}}\fi
\ifx \doiurl  \undefined \def \doiurl#1{\url{https://doi.org/#1}}\fi
\ifx \betal  \undefined \def \betal{\textit{et al.}}\fi
\ifx \binstitute  \undefined \def \binstitute#1{#1}\fi
\ifx \binstitutionaled  \undefined \def \binstitutionaled#1{#1}\fi
\ifx \bctitle  \undefined \def \bctitle#1{#1}\fi
\ifx \beditor  \undefined \def \beditor#1{#1}\fi
\ifx \bpublisher  \undefined \def \bpublisher#1{#1}\fi
\ifx \bbtitle  \undefined \def \bbtitle#1{#1}\fi
\ifx \bedition  \undefined \def \bedition#1{#1}\fi
\ifx \bseriesno  \undefined \def \bseriesno#1{#1}\fi
\ifx \blocation  \undefined \def \blocation#1{#1}\fi
\ifx \bsertitle  \undefined \def \bsertitle#1{#1}\fi
\ifx \bsnm \undefined \def \bsnm#1{#1}\fi
\ifx \bsuffix \undefined \def \bsuffix#1{#1}\fi
\ifx \bparticle \undefined \def \bparticle#1{#1}\fi
\ifx \barticle \undefined \def \barticle#1{#1}\fi
\bibcommenthead
\ifx \bconfdate \undefined \def \bconfdate #1{#1}\fi
\ifx \botherref \undefined \def \botherref #1{#1}\fi
\ifx \url \undefined \def \url#1{\textsf{#1}}\fi
\ifx \bchapter \undefined \def \bchapter#1{#1}\fi
\ifx \bbook \undefined \def \bbook#1{#1}\fi
\ifx \bcomment \undefined \def \bcomment#1{#1}\fi
\ifx \oauthor \undefined \def \oauthor#1{#1}\fi
\ifx \citeauthoryear \undefined \def \citeauthoryear#1{#1}\fi
\ifx \endbibitem  \undefined \def \endbibitem {}\fi
\ifx \bconflocation  \undefined \def \bconflocation#1{#1}\fi
\ifx \arxivurl  \undefined \def \arxivurl#1{\textsf{#1}}\fi
\csname PreBibitemsHook\endcsname

\bibitem[\protect\citeauthoryear{Ramsay and Silverman}{2002}]{Ramsay2002}
\begin{bbook}
\bauthor{\bsnm{Ramsay}, \binits{J.O.}},
\bauthor{\bsnm{Silverman}, \binits{B.W.}}:
\bbtitle{Applied Functional Data Analysis: Methods and Case Studies}.
\bpublisher{Springer},
\blocation{New York}
(\byear{2002}).
\doiurl{10.1007/b98886}
\end{bbook}
\endbibitem

\bibitem[\protect\citeauthoryear{Ferraty and Vieu}{2003}]{Ferraty2003}
\begin{barticle}
\bauthor{\bsnm{Ferraty}, \binits{F.}},
\bauthor{\bsnm{Vieu}, \binits{P.}}:
\batitle{Curves discrimination: a nonparametric functional approach}.
\bjtitle{Computational Statistics {\&} Data Analysis}
\bvolume{44}(\bissue{1-2}),
\bfpage{161}--\blpage{173}
(\byear{2003})
\doiurl{10.1016/s0167-9473(03)00032-x}
\end{barticle}
\endbibitem

\bibitem[\protect\citeauthoryear{Ramsay and Silverman}{2005}]{Ramsay2005}
\begin{bbook}
\bauthor{\bsnm{Ramsay}, \binits{J.}},
\bauthor{\bsnm{Silverman}, \binits{B.}}:
\bbtitle{Functional Data Analysis, 2nd Edn}.
\bpublisher{Springer},
\blocation{New York}
(\byear{2005}).
\doiurl{10.1007/b98888}
\end{bbook}
\endbibitem

\bibitem[\protect\citeauthoryear{Ferraty}{2011}]{Ferraty2011}
\begin{bbook}
\bauthor{\bsnm{Ferraty}, \binits{F.}}:
\bbtitle{Recent Advances in Functional Data Analysis and Related Topics}.
\bpublisher{Physica-Verlag {HD}},
\blocation{Berlin}
(\byear{2011}).
\doiurl{10.1007/978-3-7908-2736-1}
\end{bbook}
\endbibitem

\bibitem[\protect\citeauthoryear{Cuevas}{2014}]{Cuevas2014}
\begin{barticle}
\bauthor{\bsnm{Cuevas}, \binits{A.}}:
\batitle{A partial overview of the theory of statistics with functional data}.
\bjtitle{Journal of Statistical Planning and Inference}
\bvolume{147},
\bfpage{1}--\blpage{23}
(\byear{2014})
\doiurl{10.1016/j.jspi.2013.04.002}
\end{barticle}
\endbibitem

\bibitem[\protect\citeauthoryear{Febrero-Bande and de~la Fuente}{2012}]{Febrero2012}
\begin{barticle}
\bauthor{\bsnm{Febrero-Bande}, \binits{M.}},
\bauthor{\bsnm{Fuente}, \binits{M.O.}}:
\batitle{{Statistical computing in functional data analysis: The R package fda.usc}}.
\bjtitle{Journal of Statistical Software}
(\byear{2012})
\doiurl{10.18637/jss.v051.i04}
\end{barticle}
\endbibitem

\bibitem[\protect\citeauthoryear{Preda et~al.}{2007}]{Preda_2007}
\begin{barticle}
\bauthor{\bsnm{Preda}, \binits{C.}},
\bauthor{\bsnm{Saporta}, \binits{G.}},
\bauthor{\bsnm{L{\'{e}}v{\'{e}}der}, \binits{C.}}:
\batitle{{PLS} classification of functional data}.
\bjtitle{Computational Statistics}
\bvolume{22}(\bissue{2}),
\bfpage{223}--\blpage{235}
(\byear{2007})
\doiurl{10.1007/s00180-007-0041-4}
\end{barticle}
\endbibitem

\bibitem[\protect\citeauthoryear{Maturo and Porreca}{2022}]{maturo2022augmented}
\begin{barticle}
\bauthor{\bsnm{Maturo}, \binits{F.}},
\bauthor{\bsnm{Porreca}, \binits{A.}}:
\batitle{Augmented functional analysis of variance (a-fanova): theory and application to google trends for detecting differences in abortion drugs queries}.
\bjtitle{Big Data Research}
\bvolume{30},
\bfpage{100354}
(\byear{2022})
\doiurl{10.1016/j.bdr.2022.100354}
\end{barticle}
\endbibitem

\bibitem[\protect\citeauthoryear{Aguilera and Aguilera-Morillo}{2013}]{Aguilera2013}
\begin{barticle}
\bauthor{\bsnm{Aguilera}, \binits{A.}},
\bauthor{\bsnm{Aguilera-Morillo}, \binits{M.}}:
\batitle{Penalized pca approaches for b-spline expansions of smooth functional data}.
\bjtitle{Applied Mathematics and Computation}
(\byear{2013})
\doiurl{10.1016/j.amc.2013.02.009}
\end{barticle}
\endbibitem

\bibitem[\protect\citeauthoryear{Cuevas}{2014}]{Cuevas2014b}
\begin{barticle}
\bauthor{\bsnm{Cuevas}, \binits{A.}}:
\batitle{A partial overview of the theory of statistics with functional data}.
\bjtitle{Journal of Statistical Planning and Inference}
(\byear{2014})
\doiurl{10.1016/j.jspi.2013.04.002}
\end{barticle}
\endbibitem

\bibitem[\protect\citeauthoryear{Escabias et~al.}{2014}]{Escabias2014}
\begin{barticle}
\bauthor{\bsnm{Escabias}, \binits{M.}},
\bauthor{\bsnm{Aguilera}, \binits{A.M.}},
\bauthor{\bsnm{Aguilera-Morillo}, \binits{M.C.}}:
\batitle{Functional {PCA} and base-line logit models}.
\bjtitle{Journal of Classification}
\bvolume{31}(\bissue{3}),
\bfpage{296}--\blpage{324}
(\byear{2014})
\doiurl{10.1007/s00357-014-9162-y}
\end{barticle}
\endbibitem

\bibitem[\protect\citeauthoryear{Yu and Lambert}{1999}]{Yu_1999}
\begin{barticle}
\bauthor{\bsnm{Yu}, \binits{Y.}},
\bauthor{\bsnm{Lambert}, \binits{D.}}:
\batitle{Fitting trees to functional data, with an application to time-of-day patterns}.
\bjtitle{Journal of Computational and Graphical Statistics}
\bvolume{8}(\bissue{4}),
\bfpage{749}--\blpage{762}
(\byear{1999})
\doiurl{10.1080/10618600.1999.10474847}
\end{barticle}
\endbibitem

\bibitem[\protect\citeauthoryear{Gregorutti et~al.}{2015}]{Gregorutti_2015}
\begin{barticle}
\bauthor{\bsnm{Gregorutti}, \binits{B.}},
\bauthor{\bsnm{Michel}, \binits{B.}},
\bauthor{\bsnm{Saint-Pierre}, \binits{P.}}:
\batitle{Grouped variable importance with random forests and application to multiple functional data analysis}.
\bjtitle{Computational Statistics {\&} Data Analysis}
\bvolume{90},
\bfpage{15}--\blpage{35}
(\byear{2015})
\doiurl{10.1016/j.csda.2015.04.002}
\end{barticle}
\endbibitem

\bibitem[\protect\citeauthoryear{M{\"{o}}ller et~al.}{2016}]{Moller2016}
\begin{barticle}
\bauthor{\bsnm{M{\"{o}}ller}, \binits{A.}},
\bauthor{\bsnm{Tutz}, \binits{G.}},
\bauthor{\bsnm{Gertheiss}, \binits{J.}}:
\batitle{{Random forests for functional covariates}}.
\bjtitle{Journal of Chemometrics}
(\byear{2016})
\doiurl{10.1002/cem.2849}
\end{barticle}
\endbibitem

\bibitem[\protect\citeauthoryear{Rahman et~al.}{2019}]{Rahman_2019}
\begin{botherref}
\oauthor{\bsnm{Rahman}, \binits{R.}},
\oauthor{\bsnm{Dhruba}, \binits{S.}},
\oauthor{\bsnm{Ghosh}, \binits{S.}},
\oauthor{\bsnm{Pal}, \binits{R.}}:
Functional random forest with applications in dose-response predictions.
Scientific Reports
\textbf{9}(1)
(2019)
\doiurl{10.1038/s41598-018-38231-w}
\end{botherref}
\endbibitem

\bibitem[\protect\citeauthoryear{Maturo and Verde}{2023}]{maturo2023supervised}
\begin{barticle}
\bauthor{\bsnm{Maturo}, \binits{F.}},
\bauthor{\bsnm{Verde}, \binits{R.}}:
\batitle{Supervised classification of curves via a combined use of functional data analysis and tree-based methods}.
\bjtitle{Computational Statistics}
\bvolume{38}(\bissue{1}),
\bfpage{419}--\blpage{459}
(\byear{2023})
\doiurl{10.1007/s00180-022-01236-1}
\end{barticle}
\endbibitem

\bibitem[\protect\citeauthoryear{Maturo and Verde}{2022a}]{maturo2022combining}
\begin{botherref}
\oauthor{\bsnm{Maturo}, \binits{F.}},
\oauthor{\bsnm{Verde}, \binits{R.}}:
Combining unsupervised and supervised learning techniques for enhancing the performance of functional data classifiers.
Computational Statistics,
1--32
(2022)
\doiurl{10.1007/b98886}
\end{botherref}
\endbibitem

\bibitem[\protect\citeauthoryear{Maturo and Verde}{2022b}]{maturo2022pooling}
\begin{barticle}
\bauthor{\bsnm{Maturo}, \binits{F.}},
\bauthor{\bsnm{Verde}, \binits{R.}}:
\batitle{Pooling random forest and functional data analysis for biomedical signals supervised classification: Theory and application to electrocardiogram data}.
\bjtitle{Statistics in Medicine}
\bvolume{41}(\bissue{12}),
\bfpage{2247}--\blpage{2275}
(\byear{2022})
\doiurl{10.1002/sim.9353}
\end{barticle}
\endbibitem

\bibitem[\protect\citeauthoryear{Moindji{\'e} et~al.}{2024}]{moindjie2024classification}
\begin{barticle}
\bauthor{\bsnm{Moindji{\'e}}, \binits{I.-A.}},
\bauthor{\bsnm{Dabo-Niang}, \binits{S.}},
\bauthor{\bsnm{Preda}, \binits{C.}}:
\batitle{Classification of multivariate functional data on different domains with partial least squares approaches}.
\bjtitle{Statistics and Computing}
\bvolume{34}(\bissue{1}),
\bfpage{5}
(\byear{2024})
\end{barticle}
\endbibitem

\bibitem[\protect\citeauthoryear{Brault et~al.}{2024}]{brault2024mixture}
\begin{barticle}
\bauthor{\bsnm{Brault}, \binits{V.}},
\bauthor{\bsnm{Devijver}, \binits{E.}},
\bauthor{\bsnm{Laclau}, \binits{C.}}:
\batitle{Mixture of segmentation for heterogeneous functional data}.
\bjtitle{Electronic Journal of Statistics}
\bvolume{18}(\bissue{2}),
\bfpage{3729}--\blpage{3773}
(\byear{2024})
\end{barticle}
\endbibitem

\bibitem[\protect\citeauthoryear{Riccio et~al.}{2024a}]{riccio2024supervised}
\begin{botherref}
\oauthor{\bsnm{Riccio}, \binits{D.}},
\oauthor{\bsnm{Maturo}, \binits{F.}},
\oauthor{\bsnm{Romano}, \binits{E.}}:
Supervised learning via ensembles of diverse functional representations: the functional voting classifier.
arXiv preprint arXiv:2403.15778
(2024)
\end{botherref}
\endbibitem

\bibitem[\protect\citeauthoryear{Riccio et~al.}{2024b}]{riccio2024randomized}
\begin{botherref}
\oauthor{\bsnm{Riccio}, \binits{D.}},
\oauthor{\bsnm{Maturo}, \binits{F.}},
\oauthor{\bsnm{Romano}, \binits{E.}}:
Randomized spline trees for functional data classification: Theory and application to environmental time series.
arXiv preprint arXiv:2409.07879
(2024)
\end{botherref}
\endbibitem

\bibitem[\protect\citeauthoryear{Ramsay et~al.}{2009}]{Ramsay2009}
\begin{bchapter}
\bauthor{\bsnm{Ramsay}, \binits{J.}},
\bauthor{\bsnm{Hooker}, \binits{G.}},
\bauthor{\bsnm{Graves}, \binits{S.}}:
\bctitle{Introduction to functional data analysis}.
In: \bbtitle{Functional Data Analysis with R and {MATLAB}},
pp. \bfpage{1}--\blpage{19}.
\bpublisher{Springer}, \blocation{???}
(\byear{2009}).
\doiurl{10.1007/978-0-387-98185-7\_1}
\end{bchapter}
\endbibitem

\bibitem[\protect\citeauthoryear{Bagnall et~al.}{2018}]{tsc_repository}
\begin{botherref}
\oauthor{\bsnm{Bagnall}, \binits{A.}},
\oauthor{\bsnm{Lines}, \binits{J.}},
\oauthor{\bsnm{Bostrom}, \binits{A.}},
\oauthor{\bsnm{Large}, \binits{J.}},
\oauthor{\bsnm{Keogh}, \binits{E.}}:
The UEA \& UCR Time Series Classification Repository.
Accessed: 2024-09-08
(2018).
\url{http://www.timeseriesclassification.com}
\end{botherref}
\endbibitem

\bibitem[\protect\citeauthoryear{Thakoor and Gao}{2005}]{Thakoor2005}
\begin{bchapter}
\bauthor{\bsnm{Thakoor}, \binits{N.}},
\bauthor{\bsnm{Gao}, \binits{J.}}:
\bctitle{Shape classifier based on generalized probabilistic descent method with hidden markov descriptor}.
In: \bbtitle{Tenth IEEE International Conference on Computer Vision (ICCV'05) Volume 1},
vol. \bseriesno{1},
pp. \bfpage{708}--\blpage{713}
(\byear{2005}).
\doiurl{10.1109/ICCV.2005.52}
\end{bchapter}
\endbibitem

\bibitem[\protect\citeauthoryear{Cuevas et~al.}{2007}]{Cuevas_2007}
\begin{barticle}
\bauthor{\bsnm{Cuevas}, \binits{A.}},
\bauthor{\bsnm{Febrero}, \binits{M.}},
\bauthor{\bsnm{Fraiman}, \binits{R.}}:
\batitle{Robust estimation and classification for functional data via projection-based depth notions}.
\bjtitle{Computational Statistics}
\bvolume{22}(\bissue{3}),
\bfpage{481}--\blpage{496}
(\byear{2007})
\doiurl{10.1007/s00180-007-0053-0}
\end{barticle}
\endbibitem

\bibitem[\protect\citeauthoryear{{Taiwo Ojo} et~al.}{2021}]{fdaoutlier}
\begin{botherref}
\oauthor{\bsnm{{Taiwo Ojo}}, \binits{O.}},
\oauthor{\bsnm{Lillo}, \binits{R.}},
\oauthor{\bsnm{{Fernandez Anta}}, \binits{A.}}:
Fdaoutlier: Outlier Detection Tools for Functional Data Analysis.
(2021).
R package version 0.2.0.
\url{https://CRAN.R-project.org/package=fdaoutlier}
\end{botherref}
\endbibitem

\bibitem[\protect\citeauthoryear{Conn et~al.}{2022}]{fuzzyforest}
\begin{botherref}
\oauthor{\bsnm{Conn}, \binits{D.}},
\oauthor{\bsnm{Ngun}, \binits{T.}},
\oauthor{\bsnm{Ramirez}, \binits{C.M.}}:
Fuzzyforest: Fuzzy Forests for Feature Selection in the Presence of Correlated Covariates.
(2022).
\doiurl{10.18637/jss.v091.i09} .
R package version 1.0.8.
\url{https://cran.r-project.org/web/packages/fuzzyforest/fuzzyforest.pdf}
\end{botherref}
\endbibitem

\end{thebibliography}

\end{document}